\documentclass[11pt]{article}

\usepackage[T1]{fontenc}
\usepackage[utf8]{inputenc}
\usepackage[margin=1in]{geometry}
\usepackage{microtype}
\usepackage{amsmath,amssymb}
\usepackage{booktabs}
\usepackage{array}
\usepackage{tabularx}
\usepackage{longtable}
\usepackage{enumitem}
\usepackage{graphicx}
\usepackage{xcolor}
\usepackage{hyperref}
\usepackage{caption}
\usepackage{subcaption}
\usepackage{float}

\newcommand{\PaperTableStretch}{1.22}
\renewcommand{\arraystretch}{\PaperTableStretch}
\newcommand{\PaperStack}[2]{\shortstack[l]{#1\\#2}}
\newcommand{\PaperStackHead}[2]{\shortstack[c]{#1\\#2}}
\newcolumntype{Y}{>{\raggedright\arraybackslash}X}
\newif\ifshowclaimmanifest
\showclaimmanifesttrue

\definecolor{accentBlue}{HTML}{1D3557}
\definecolor{accentGold}{HTML}{B08900}

\hypersetup{
  colorlinks=true,
  linkcolor=accentBlue,
  citecolor=accentBlue,
  urlcolor=accentBlue
}

\IfFileExists{generated/tex/build_meta.tex}{\newcommand{\PrimaryRepoSHA}{7f1c37a}

\newcommand{\PrimaryRepoCommitDate}{2026-05-05}

}{
  \newcommand{\PrimaryRepoSHA}{metadata unavailable}
  
  \newcommand{\PrimaryRepoCommitDate}{metadata unavailable}

}
\newcommand{\DraftDate}{\PrimaryRepoCommitDate}

\title{VLMaxxing through FrameMogging\thanks{Look, they said attention was all you needed, well we got your attention with this title.}\\Training-Free Anti-Recomputation for Video Vision-Language Models}
\author{
  JF Bastien\\
  \texttt{paper@jfbastien.com}
  \and
  Sam D'Amico\\
  \texttt{sam@impulselabs.com}
}
\date{\DraftDate}

\begin{document}
\maketitle

\begin{abstract}
Video vision-language models (VLMs) keep paying for visual state the stream
already told us was stable. The factory wall did not move, the stream has cheap
evidence that it did not move, and the runtime still hands the model dense RGB
or a fresh prefix again. We study that waste as training-free
anti-recomputation: reuse state when validation says it survives, and buy fresh
visual evidence when the scene, query, or cache topology requires it.

The biggest measured win comes after ingest. On frozen
Qwen2.5-VL-7B-Instruct-4bit, adaptive same-video follow-up reuse preserves
paired choices and correctness on the 93-query VideoMME breadth setting while
reducing follow-up latency by 14.90--35.92$\times$. The first query is still
cold; the win starts when later questions reuse the same video state.
Selective re-prefill moves reused state out of the cache basin, and the second
follow-up inherits that repaired cache instead of buying the visual tail again.
Stress tests bound
the result rather than pretending it is universal: adaptive and scheduled-refresh
policies are drift-clean through a 50-turn repeated-question stress, fixed
\(K=1\) stays below the gate with sparse drift, and dense-answer-anchored prompt
variation separates conservative repair from faster aggressive policies that
drift.

Fresh-video pruning is smaller but real. First-pass vision pruning
(C-VISION) skips timed vision-tower work before the first answer is generated.
On Gemma 4-E4B-4bit, the clean 32f short cell reaches 1.316$\times$
first-query speedup with no paired drift or parse failures on 20 items. Qwen
provides the cautionary half: an 8f keep-rate sweep follows the timing ceiling
while failing fidelity, and a conservative 16f point recovers to within the
aggregate-accuracy/format gates only at low gain.

The accounting is the point. Stage-share ceiling (C-CEILING) says a component
speedup becomes an end-to-end speedup only in proportion to the wall-clock share
it actually accelerates. That explains why first-pass gains are modest, why the
n=60 composition audit lands near the measured ceiling, and why C-VISION and
after-ingest follow-up reuse do not multiply just because both numbers are in
the paper.

Streaming state reuse (candidate C-STREAM) is the native-rate deployment target,
not a headline result here. The 26B-class scale-out bundle shows the gates:
default cross-turn cache reuse can be unsafe, after-warm prefix snapshots can
help at small \(N\), and low-FPS dense remains a serious fixed-evidence
baseline. The larger direction is that future VLM-native media should expose
change, motion, uncertainty, object state, sensor time, and active tiles
directly, so models do not have to rediscover the world from dense RGB every
frame.

\end{abstract}

\begin{figure}[H]
  \centering
  \includegraphics[width=\linewidth]{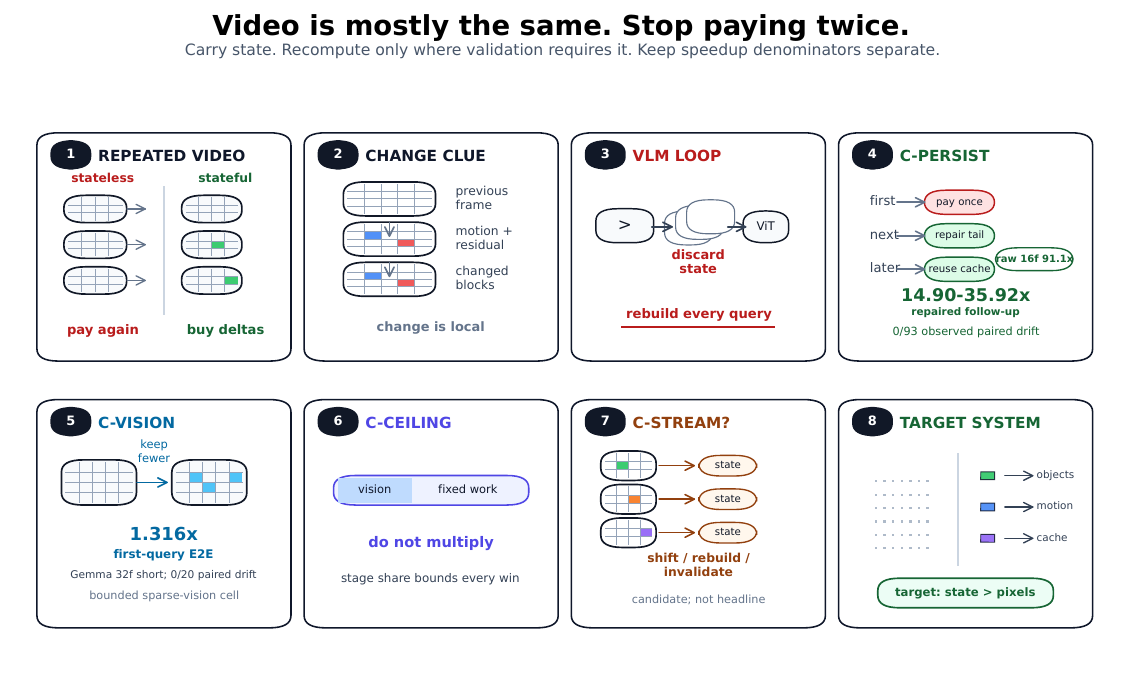}
  \caption{Graphical overview. Video state is reused when validation gates
  allow it and refreshed when the scene, query, or cache topology requires it.
  C-PERSIST, C-CEILING, and C-VISION are the main regimes; candidate C-STREAM
  remains a state-update target.}
  \label{fig:regime-overview}
\end{figure}

\section{Introduction}

Video vision-language models (VLMs) spend a surprising amount of time
rediscovering what they already know. The wall behind a factory robot did not
move. The stove, countertop, or
security-camera background often did not move either. A vehicle's front camera
also sees much of the world move predictably: nearby objects expand, lane
markers flow outward, and the surprise is concentrated near boundaries,
occlusions, and independent movers. The video stream carries cheap evidence of
stasis, motion, and residual surprise, and classical codecs have spent decades
turning that evidence into compact state. Yet most VLM pipelines hand the model
another dense stack of RGB frames and ask it to pay again for vision encoding,
prefill, or both.

That repeated payment is the seed of this paper. A benchmark clip is sampled
into several frames, many of those frames are visually redundant, and the stack
still treats them as fresh. The same waste appears in repeated follow-up
questions on the same video, and it becomes even more obvious in live streams
where most of the scene remains stable for long stretches. The interesting
question is no longer whether redundancy exists. It does. The question is
where that redundancy can be removed without breaking the answer.
The largest measured opportunity is after ingest: repaired same-video
follow-up reuse reaches 14.90--35.92$\times$ speedup with no observed paired
drift across 93 paired queries, including 62 repaired follow-up turns.

Our central question is where a VLM must buy fresh visual evidence over time,
and where cached state is enough to stay on the relevant regime-specific
quality--compute frontier. The key variable is not simply how much visual
compute is reused. A small transient event, a moving hand, a text flash, or an
object entering at a boundary can matter more than a large region of
predictable background motion. The planner buys fresh evidence; the model
still decides what matters.

The literature often treats that waste as if it were one knob. It is not. The
fresh-video pass is bounded by whatever share of wall-clock the vision tower
actually owns. After-ingest follow-up queries on the same video are a
different regime: once the prefix has been paid, prompt reuse can dominate.
Routing under a fixed dense backend is different again: it tells us whether a
training-free policy preserves answer quality, not whether the backend has
already been rewritten to skip timed work. Reuse is three regimes with three
different ceilings, not one multiplier that can be carried from table to table.

That distinction changes what the paper is about. We make three contributions.
\emph{Persistent follow-up reuse (C-PERSIST)} repairs same-video after-ingest
reuse, where the biggest multipliers live but where unrepaired caches can fall
into a measurable basin. \emph{Stage-share ceiling (C-CEILING)} is the
arithmetic guardrail: a local speedup survives to end-to-end wall-clock only in
proportion to the stage it actually accelerates. \emph{First-pass vision
pruning (C-VISION)} measures training-free vision-tower work skipped on fresh
videos. Streaming remains the native-rate target and boundary evidence, not a
fourth headline; Qwen routing remains mechanism evidence under a fixed dense
backend.

The larger target is not a bigger threshold sweep and not a new codec in this
paper. The experiments are a measurement instrument: they ask which
cheap temporal signals are already useful to a frozen VLM runtime, where those
signals fail, and what a future VLM-native media stream should expose directly.
Codec and pixel signals are therefore treated as freshness and uncertainty
signals, not as semantic-saliency oracles. They can say where something
probably changed; the model and the task still decide what matters. That is
why the VLM-native media discussion later in the paper is framed as candidate
requirements, not as a systems claim here.

The scale-out streaming lane is part of the same science story, but it has a
different protocol. A 26B-class bundle sharpens the boundary rather than simply
adding wins: the default cross-turn cache path is unsafe, topology-aware
after-warm prefix snapshots are positive at small \(N\), and fixed-evidence
streaming baselines show that low-FPS dense is a serious competitor. The same warning
shows up locally in Qwen session-composition experiments, where speedup can
survive while the full follow-up fidelity contract does not.

\begin{table}[H]
\centering
\caption{Reuse regime map. Rows are separate workloads and denominators, not
one multiplier.}
\label{tab:three-regimes}
\footnotesize
\renewcommand{\arraystretch}{\PaperTableStretch}
\begin{tabularx}{\linewidth}{@{}
  >{\raggedright\arraybackslash}p{0.18\linewidth}
  >{\raggedright\arraybackslash}p{0.23\linewidth}
  >{\raggedright\arraybackslash}p{0.22\linewidth}
  >{\raggedright\arraybackslash}X
@{}}
\toprule
Regime & Reused work & Denominator / gate & Evidence / boundary \\
\midrule
After-ingest follow-up & prompt / key--value (KV) prefix for later questions on the same video & median cold follow-up / cached-or-repaired follow-up plus paired drift; first query still pays full freight & 0/93 paired drift in breadth protocol at 14.90--35.92$\times$; stress tests expose conservative/aggressive boundary \\
Fresh-video first pass & vision-tower work inside the first query & first-query wall-clock; clean/advisory gates plus measured sparse-execution envelopes & VideoMME clean 1.113$\times$; Gemma 32f short 1.316$\times$ timed skip; Qwen low-gain boundary \\
Candidate C-STREAM & repeated visual state in a native-rate or live pipeline & after-warm prefix snapshot timing plus fixed-evidence baselines; not setup-inclusive throughput & default 26B cache path blocked; after-warm prefix snapshots positive but small-\(N\); low-FPS dense is a strong baseline \\
\bottomrule
\end{tabularx}
\end{table}

The table is the denominator map. The contributions below explain those
workloads without collapsing them into a single multiplier:
\begin{itemize}[leftmargin=1.5em]
\item \textbf{C-PERSIST:} same-video follow-up reuse with selective visual-tail
repair. Adaptive repaired-cache inheritance preserves paired choices and
correctness on the 93-query breadth setting and reaches 14.90--35.92$\times$
follow-up speedup; stress tests expose the conservative/aggressive boundary.
\item \textbf{C-CEILING:} a denominator guardrail. The same arithmetic explains
both modest first-pass multipliers and the n=60 composition audit, where the
observed 1.042$\times$ stacked result lands where measured share terms predict.
\item \textbf{C-VISION:} measured first-pass vision pruning. Gemma gives a
clean 1.316$\times$ timed-skip cell at 32f short; Qwen shows the
fidelity/speed boundary rather than a universal sparse-backend claim.
\end{itemize}

Those three contributions do not compose as a naive product. C-VISION changes
first-pass latency. C-PERSIST changes follow-up latency on later queries about
the same video. C-CEILING says shared stage budgets cannot be spent twice. The
paper therefore reports composition against measured ceilings, not against
headline multiplication.

Qwen routing keeps that framing honest. It shows a quality--compute frontier
under a frozen dense backend, not a timed sparse implementation. The same
discipline applies to paired drift: several runs preserve aggregate accuracy
while changing individual answers, so answer identity, correctness, and format
are reported separately whenever they decide the claim boundary.

\section{Related Work}

The closest systems neighbor is CodecSight \cite{codecsight}, which also tries
to avoid unnecessary visual processing by using compressed-video structure
before the rest of the model consumes the content. The difference is both in
signal and in target. CodecSight uses decoder-native motion vectors with
optional residuals; its default NVDEC runtime uses motion vectors because
residuals are not exposed. It is evaluated on streaming anomaly detection; our
local implementation uses a pixel-domain proxy because that is what our
MLX stack can exercise today, and our main benchmark evaluations target
temporal-reasoning QA on TOMATO \cite{tomato} and MVBench \cite{mvbench}.

Trained codec-native approaches answer a different question. CoPE-VideoLM
\cite{copevideolm}, Deja Vu \cite{dejavu}, and older compressed-video pipelines
such as CoViAR \cite{coviar} learn how motion, residual, or reused
representations should enter the model. That is valuable prior art, but it does
not settle the deployment question we care about here: how much redundancy a
frozen stack can exploit before retraining or representation learning enters
the picture.

Another family reduces visual cost after tokens already exist. FastV
\cite{fastv}, FastVID \cite{fastvid}, FlashVID \cite{flashvid}, VisionZip
\cite{visionzip}, FrameFusion \cite{framefusion}, SparseVLM
\cite{sparsevlm}, VScan \cite{vscan}, STTM
\cite{sttm}, SparseVILA \cite{sparsevila}, and LLaVA-PruMerge
\cite{llavaprumerge} prune, retrieve, or merge visual tokens using
attention-derived, density, query-aware, or feature-similarity signals. ToMe
\cite{tome} is the older training-free Vision Transformer (ViT)
token-merging ancestor, and
EvoPrune \cite{evoprune} is a late-breaking MLLM example of pruning inside
the visual encoder itself. Those methods are adjacent but not identical to our
setting. Their signal usually lives after visual tokens exist, or changes the
encoder/runtime contract. Our routing lane asks an earlier question: can we
decide where fresh computation should land in time before the full dense
backend pays for every frame again? Our first-pass pruning lane then adds a
different discipline: the speedup must be interpreted through the measured
stage-share ceiling, not through the component multiplier alone.

Temporal redundancy inside vision transformers is also prior art. Eventful
Transformers \cite{eventful} reprocess only tokens that changed significantly
between frames, often without retraining. That makes it a useful north star for
measured sparse execution. Our local measured sparse-execution evidence is
deliberately narrower: it verifies timed vision-work reduction, shows a clean
Gemma operating point, and reports Qwen's keep-rate boundary instead of
claiming a broad sparse backend.

Long-horizon reuse and systems co-design are directly relevant. QuickVideo
\cite{quickvideo} explicitly co-designs decode and prefill. ReKV \cite{rekv},
VLCache \cite{vlcache}, StreamingVLM \cite{streamingvlm}, SparseVILA
\cite{sparsevila}, and HERMES \cite{hermes} study cached visual or KV state
over long contexts, streaming video, or multi-turn multimodal inference.
SGLang's RadixAttention \cite{sglang}, vLLM/PagedAttention \cite{vllm}, and
standard prefix caching show that shared-prefix KV reuse is already a serving
systems primitive; C-PERSIST should not be read as inventing that primitive.
StreamingLLM \cite{streamingllm} is also adjacent as a text-only long-context
KV-retention result built around attention sinks, though it is not a visual
state-reuse method.
CacheBlend \cite{cacheblend} is even closer to the repair pattern: it reuses
bulk cached KV for RAG contexts and selectively recomputes part of the cache to
restore quality. Our contribution is the video-VLM instantiation and boundary:
which visual tail to refresh, the cache-basin failure mode, the paired-drift
contract, and the repaired-cache inheritance rule that separates the failed and
successful adaptive variants.

Efficient VLM architectures such as FastVLM \cite{fastvlm} attack the same
latency pressure by changing the visual encoder and tokenization. They are
complementary to this paper rather than baselines for a frozen-stack claim:
our experiments ask what a deployed VLM can avoid recomputing without changing
weights, while C-CEILING explains when even a large component win can remain a
modest request-level win.

SimpleStream \cite{simplestream} is also an important streaming baseline
because it argues that a strong recent-frame window can match or outperform
heavier streaming-memory systems on modern benchmarks. That critique matters
directly for our streaming lane: any cache or memory mechanism has to beat a
strong recency baseline under matched budget, not just beat a strawman.

More conceptually, our framing is adjacent to approximate computing
\cite{rinardapprox}: skipping work under an explicit quality budget. The
difference is that the skipped work here is temporal visual recomputation, and
the guardrail is benchmark fidelity rather than an application-specific numeric
tolerance.

Our position is therefore narrower and more explicit than a generic
``efficient-video-VLM'' claim. We study training-free anti-recomputation for
frozen video VLMs, separating first-pass reuse, after-ingest reuse, and
mechanism-validation routing. The paper is not a claim that pixel differencing
is the final signal. It is a claim that careful temporal reuse is already worth
studying before codec-native systems are available in this stack.

\section{Method}
\label{sec:method}

\subsection{Problem Setting}

We study three deployment regimes on frozen video-VLM stacks, plus a separate
routing path used for mechanism validation:

\begin{itemize}[leftmargin=1.5em]
\item \textbf{First-pass visual pruning:} skip or reduce vision-tower work
before the rest of the model pays for it.
\item \textbf{After-ingest follow-up reuse:} keep the expensive prefix from the
first query and ask the next question against the same video.
\item \textbf{Streaming / deployment reuse:} revisit repeated visual state in
a native-rate or live pipeline.
\item \textbf{Routing under a fixed dense backend:} substitute cached visual
state to test whether answer quality survives, even when the benchmark path has
not yet been rewritten into a sparse wall-clock implementation.
\end{itemize}

These are related, but they are not interchangeable. They are kept
analytically separate because they measure different kinds of reuse.

\subsection{Reader's Map of the Session Terms}

The session experiments are easier to read with the naming convention stated
up front. A \emph{cold} query recomputes the full video prefix. A
\emph{follow-up} query asks another question about the same video after that
prefix has already been ingested. The named first query creates reusable
state; later same-video questions are follow-ups in the same session. The
symbol \(K\) is the number of newest visual frames re-prefilled before a
follow-up; \(K=1\) means ``refresh one newest frame, reuse the rest of the
prefix.'' A \emph{same-class speedup}
compares cold follow-up queries to session follow-up queries. An
\emph{all-query denominator} compares all cold queries to session follow-ups;
we report it only when the text says so explicitly. For example, a cold median
over all three queries divided by the repaired-session follow-up median is an
all-query speedup; it does not include the repaired session's cold first query
in the repaired follow-up median or pretend the first query was free.

The main paired-fidelity metric is \emph{paired drift}. A run has no observed paired
choice drift when each session answer chooses the same multiple-choice option
as its matched cold baseline. It has no observed paired correctness drift when
each answer is right or wrong on the same examples as the baseline. This is
stronger than matching aggregate accuracy, but it is still a finite-sample
observation, not a population proof.

A \emph{cache basin} is the corresponding failure region: reused state drives
systematic answer drift or repeated pathological outputs rather than ordinary
independent mistakes. Selective re-prefill is designed to leave that basin by
refreshing a small newest-frame visual tail before a follow-up query.

\subsection{Temporal Planner}

For adjacent frames, the Qwen routing path uses max-absolute RGB
difference (\texttt{max\_abs}) over model-aligned blocks; other block
statistics remain implementation options. Fixed
thresholds partition blocks into \emph{static}, \emph{shifted}, and
\emph{novel} regions. These names are historical planner labels, not literal
codec semantics: today they mean low, medium, and high pixel-domain change.

The block geometry is derived from the model's vision configuration rather than
hard-coded. On Qwen2.5-VL, planner decisions line up with the merged-token
spatial grid so cached features can be substituted back into the dense path at
the same geometry the benchmark runner already understands.

The paper's default sparse-QA routing path still uses pixel differences because
that is the validated semantic baseline. A codec-native planner bridge remains
semantic substitution evidence, not a latency claim: replacing the pixel-diff
planner signal with H.264-derived scores can preserve local dense choices, but
a systems claim requires a streaming decoder-integrated path.

This distinction is load-bearing. Pixel differences, motion vectors, residual
proxies, and packet structure are novelty or uncertainty signals, not
semantic-saliency signals. They help decide where fresh visual evidence should
be bought; they do not decide whether a changed region is task-relevant.
Appendix~\ref{app:real-clip-planner} visualizes this routing planner on real
clips, separating class-overlaid audit views from readable reused/fresh budget
summaries.

\subsection{Bounded Staleness}

Reuse without age control drifts. The planner therefore carries a
bounded-staleness counter: even a block that continues to look reusable must be
refreshed after its age exceeds a configured limit. The reported policy sets
that limit to four frame steps. That choice is not cosmetic.
It is an important guardrail on TOMATO, where temporally concentrated evidence
punishes policies that keep trusting a stale region simply because the local
pixel difference remains small.

We summarize routing budget by converting block-level reuse into an effective
fresh-frame count,
\[
f_{\mathrm{eff}} = 1 + (N - 1)(1 - r_{\mathrm{reuse}}),
\]
where $r_{\mathrm{reuse}}$ is the mean active-region reuse ratio. This gives
the Qwen routing results a common x-axis with dense-$N$ baselines: how much
fresh visual budget the policy effectively spent.

\subsection{The Frontier We Measure}

The paper's scientific object is not raw reuse rate. It is the
quality--compute frontier produced by placing fresh evidence in time. In the
Qwen routing lane, the x-axis is effective fresh frames and the y-axis is
answer quality or dense agreement. In the first-pass lane, the x-axis is
measured wall-clock after vision-tower pruning and the y-axis is benchmark
accuracy. In the follow-up lane, the x-axis is session follow-up latency after
the first query has paid full freight, and the y-axis is paired drift against
cold follow-up answers.

Those are separate frontiers, not one universal curve. A semantic-substitution
point, where cached visual features are substituted into the dense runner to
test answer preservation, can show that a routing policy preserves answers at
lower effective fresh-frame budget; it becomes a systems speedup only after
the backend skips timed work. A persistent-cache point can show large
same-video follow-up gains; it says nothing about first-pass latency for a
fresh video. The manuscript
therefore reports quality at fixed budget, budget at fixed quality, and the
elbow or boundary point inside each regime, rather than multiplying the best
number from each table.

\subsection{Vision-Tower Pruning and Persistent-KV}

The other two mechanisms operate at different points in the stack.

The Gemma path patches the vision tower after an internal layer, keeps only a
fraction of the tokens, and scatters the result back so downstream geometry is
unchanged. \(L\) names the internal pruning layer, and \(kr_V\) is the keep
ratio for visual token groups after that layer; \(kr_{Q0}\) is the same keep
ratio when applied only to the first-query admission pass in the Qwen bridge;
\(K\) is reserved for newest-frame re-prefill in C-PERSIST. The resulting first-pass
speedup is constrained by how much of total wall-clock the vision tower owns.
More generally, if a method accelerates only
one stage of a pipeline by a factor $s$, and all other stages are held fixed,
the ideal end-to-end gain is
\[
\mathrm{E2E}_{\mathrm{ideal}} =
\frac{1}{f_{\mathrm{fixed}} + (1 - f_{\mathrm{fixed}}) / s},
\]
where $f_{\mathrm{fixed}}$ is the dense wall-clock fraction left untouched.
That is the paper's C-CEILING contribution: the denominator limits how much of
a component speedup can survive. In measured runs, residuals around this model
expose overhead or non-target-stage timing variation despite the intended
intervention. For vision-tower pruning, the same logic
reduces to the scatter-back prediction
\[
\mathrm{E2E}_{\mathrm{pred}} =
\frac{1}{1 - V_{\mathrm{share}} V_{\mathrm{red}}},
\]
where $V_{\mathrm{share}}$ is the dense vision share and $V_{\mathrm{red}}$ is
the observed reduction in vision time. Both are measured from stage timings:
\(V_{\mathrm{share}}\) from the dense baseline, \(V_{\mathrm{red}}\) from the
paired pruned run. The ceiling prediction is then compared with observed
end-to-end speedup; it is not fit to the observed end-to-end result.

The persistent-KV path answers a different question. Here the user has already
paid for the first query on a video. The next question about that same video
reuses the prompt cache and only appends a short textual suffix. The result can
be dramatically faster, but it is explicitly an \emph{after-ingest} claim. It
is not a claim about faster first-pass understanding of fresh videos.

Persistent reuse also has a repair path. When a long visual prefix falls into a
cache-reuse basin, the selective re-prefill runner keeps the reusable text and
old visual prefix, refreshes a fixed newest-frame visual tail, and continues the
follow-up from that repaired state. The knob \(K\) is the number of newest
visual frames re-prefilled on each follow-up. Larger \(K\) spends more latency
to repair more visual context; smaller \(K\) approaches the deployment-speed
regime. This is still after-ingest reuse: the first query remains cold, and the
claim is about later questions on the same video.

\subsection{Paired Drift Metrics}

Aggregate accuracy is not a sufficient paired-fidelity metric for reuse. A cached or
pruned run can keep the same aggregate score while changing individual answer
choices, and a follow-up cache can lose accuracy by collapsing into a
pathological output distribution rather than by making ordinary wrong choices.
The paper therefore records paired choice diffs, paired correctness diffs,
parse failures, degenerate or pathological-like outputs, and query-index splits
whenever these decide the claim. In the C-PERSIST audits, pathological-like
means the raw response matched the repeated malformed attractor strings logged
in the unrepaired basin, including \texttt{addCriterion...} or the literal
four-character Chinese string for ``auto-generated'' (Unicode
U+81EA U+52A8 U+751F U+6210).
The distinction matters scientifically:
answer-identity drift, accuracy drift, and format collapse point to different
repair mechanisms.

\subsection{What Is Actually Skipped}

The paper keeps semantic substitution and wall-clock skipping separate. Qwen
routing can replay or substitute cached features to test whether the answer
survives; it is not an end-to-end speedup claim until the backend skips timed
work. Gemma sparse vision and persistent-KV follow-up reuse do change measured
wall-clock and therefore support speedup claims. The Qwen session/streaming
bridge is Q0 admission plus cache-state reuse, not active follow-up pruning:
follow-up image tokens were cache-served.

\begin{table}[H]
\centering
\caption{Runtime hooks and validation gates required to turn each reuse
mechanism into a systems claim without confusing semantic reuse, real skipped
work, and correctness.}
\label{tab:runtime-contract}
\scriptsize
\renewcommand{\arraystretch}{\PaperTableStretch}
\begin{tabularx}{\linewidth}{@{}>{\raggedright\arraybackslash}p{0.18\linewidth} Y Y Y@{}}
\toprule
Regime & Runtime hook & Validation gate & Failure mode to log \\
\midrule
Fresh-video vision pruning & vision-tower keep masks, token geometry, per-stage
timing & paired accuracy/format, \(V_{\mathrm{share}}\), \(V_{\mathrm{red}}\),
and ceiling residual & format collapse, low \(V_{\mathrm{share}}\), policy
overhead \\
After-ingest follow-up reuse & stable prefix-cache semantics, newest-frame
tail re-prefill, cache-state source labels & paired choice/correctness drift,
pathological outputs, setup and follow-up latency & cache basin, wrong
inheritance source, silent cache-topology corruption \\
Candidate streaming state reuse & stream-time staleness signals, shift/rebuild
events, raw source media, matched simple baselines & cache-correctness check,
low-FPS/screenshot/recency comparisons, stale-cache case & stale freshness,
compute denial, proxy loses to simple dense baseline \\
Routing/control lane & frame-selection logs, novelty-ranked dense controls,
effective-fresh-frame accounting & quality frontier under dense backend; no
speed claim & confusing answer preservation with timed work skipped \\
\bottomrule
\end{tabularx}
\end{table}

\section{Experimental Setup}

\subsection{Benchmarks and Workloads}

The paper uses three headline workload families plus one scale-out boundary
lane, because the contribution is not a single benchmark trick and the
protocols are not interchangeable.

The routing mechanism is evaluated on sparse semantic-substitution holdouts:
TOMATO \cite{tomato} and a frozen motion-heavy MVBench slice \cite{mvbench}.
These are the cleanest settings for testing whether training-free temporal
reuse preserves answers at lower effective fresh-frame budgets under a fixed
dense backend.

The first-pass speedup story uses the MLX-community
\texttt{gemma-4-e4b-4bit} conversion of Google Gemma 4-E4B on VideoMME
\cite{gemma4e4b,gemma4e4bmlx,videomme}, MVBench, and TOMATO at 8-frame
settings, plus dev-expansion cells and deeper frame-count probes. The
after-ingest persistent-KV story uses repeated follow-up questions on VideoMME
sessions across short, medium, long, and deeper 32-frame slices, where the same
video is queried multiple times and prompt reuse is therefore meaningful.

The scale-out streaming and UI studies are native-rate deployment evidence
under a separate protocol, not replacements for the benchmark evaluation.
By design, sparse routing uses pixel-diff on sampled decoded frames, while the
separate streaming lane uses H.264 metadata at native frame rate, including
FFmpeg motion-vector side data and codecview inspection paths
\cite{ffmpegcodecs,ffmpegfilters}. The signals are protocol-specific, not
interchangeable.

\subsection{Models, Hardware, and Preprocessing}

The Qwen routing and persistent-KV experiments use Qwen2.5-VL-7B-Instruct-4bit
\cite{qwen25vl}, a frozen open-weight video VLM, through MLX-VLM
\cite{mlxvlm}. The first-pass vision-pruning results use the MLX-community
4-bit conversion of Gemma 4-E4B \cite{gemma4e4b,gemma4e4bmlx}, a separate open
VLM family. The scale-out streaming evaluation uses a separate RTSP and
screen-recording pipeline with a Gemma 4 26B/A4B-family stack. Table~\ref{tab:setup-lanes}
records the exact hardware for each lane.

For benchmark comparability, the main-path runs use uniform global sampling at
8 frames and square-pad plus resize to 560\,$\times$\,560. The routing planner
operates on decoded RGB frames. Padding-aware reuse accounting is reported
separately from raw padded reuse; reported effective-budget numbers use
active-region reuse rather than trivially static padded borders.

\begin{table}[H]
\centering
\caption{Main evaluation lanes. The rows are protocol identifiers, not a
claim that all results share one denominator.}
\label{tab:setup-lanes}
\small
\renewcommand{\arraystretch}{\PaperTableStretch}
\begin{tabularx}{\linewidth}{@{}>{\raggedright\arraybackslash}p{0.20\linewidth} >{\raggedright\arraybackslash}p{0.23\linewidth} Y >{\raggedright\arraybackslash}p{0.20\linewidth}@{}}
\toprule
Lane & Model/runtime & Primary protocol & Main reported metric \\
\midrule
Routing & Qwen2.5-VL-7B-Instruct-4bit, MLX-VLM / M3 Air 16\,GB & TOMATO and MVBench
semantic-substitution holdouts & Accuracy/agreement at effective fresh-frame
budget \\
First-pass pruning & mlx-community/gemma-4-e4b-4bit and matched Qwen probes, MLX / M3 Air 16\,GB & Fresh-video
first query at 8f plus selected frame-count probes & First-query E2E, \(V_{\mathrm{share}}\),
\(V_{\mathrm{red}}\), accuracy delta \\
After-ingest reuse & Qwen2.5-VL-7B-Instruct-4bit, MLX-VLM / M3 Air 16\,GB & VideoMME sessions
with one cold query and two follow-ups & Paired drift, follow-up latency,
repair speedup \\
Scale-out streaming & Gemma 4 26B/A4B-family RTSP / screen-recording stack,
M5 Max MacBook Pro 128\,GB &
Separate native-rate scale-out protocol & ViT-call reduction, throughput, diagnostics,
evidence boundary \\
\bottomrule
\end{tabularx}
\end{table}

For local timing claims, the harness uses one wall-clock source per experiment,
forces the backend to finish work before stopping the timer, and reports the
denominator for every speedup. Follow-up speedups are median cold-follow-up
latency divided by median session-follow-up latency unless explicitly labeled
otherwise; first-pass rows are first-query end-to-end ratios. Paired
experiments keep clip order and compared conditions aligned, log raw per-query
records, and label rows advisory when non-target-stage timing movement prevents
clean attribution. The practical pairing rule is simple: compared arms use the
same input list, frame count, scorer, and wall-clock harness, and the advisory
label follows the logged timing deltas. When a run is missing the
instrumentation needed for a mechanism claim, the manuscript says so or keeps
the result advisory.

\subsection{Metrics and Reporting Discipline}

We report different metrics for different regimes because the paper would be
misleading otherwise.

For Qwen routing, the paper reports accuracy, dense-versus-cached agreement,
parse failures, and effective fresh frames. For Gemma first-pass pruning, it
reports end-to-end speedup, vision-time reduction, accuracy delta, and the
share$\times$reduction prediction. For persistent-KV, it reports follow-up
latency, cold-versus-session speedup, prefix coverage, paired choice and
correctness drift, and accuracy delta between baseline and session modes. These
paired-drift metrics are stricter than aggregate accuracy: they ask whether the
same item gets the same parsed answer under reuse, not merely whether the total
score is unchanged.

For the adaptive C-PERSIST breadth result, the denominator is both query- and
session-visible. The combined envelope is 31 sessions \(\times\) three queries
per session \(=93\) paired queries, with 62 repaired follow-ups. The paper
reports the query-level no-observed-drift bound because the paired answer is
logged per query, and also reports the more conservative 0/31 session-level
view where any drift in a session would count against the session.

\section{Quality--Compute Frontiers Under Measured Ceilings}
\label{sec:results-cross-architecture}

The strongest measured results in this paper appear when the reused stage
actually owns meaningful wall-clock. That sounds obvious, but it is the part
component-level reporting can blur. C-CEILING is the paper's arithmetic sanity
check: local multipliers only survive to end-to-end wall-clock in proportion to
the share of time the accelerated stage actually owns. The gains below are
therefore not one number. They are regime-dependent by construction.

\begin{table}[H]
\centering
\caption{Main speedup cells. Rows use different denominators; C-CEILING explains why they do not multiply.}
\label{tab:headline-results}
\scriptsize
\setlength{\tabcolsep}{2pt}
\renewcommand{\arraystretch}{\PaperTableStretch}
\begin{tabularx}{\linewidth}{@{}>{\raggedright\arraybackslash}p{0.12\linewidth} Y >{\raggedright\arraybackslash}p{0.15\linewidth} >{\raggedright\arraybackslash}p{0.10\linewidth} Y >{\raggedright\arraybackslash}p{0.12\linewidth}@{}}
\toprule
Regime & Setting & Speedup denominator & Gain & Validation / notes & Evidence \\
\midrule
After-ingest repair & Qwen adaptive repair breadth & cold/repaired follow-up & 14.90--35.92$\times$ & choice 0/93; correct 0/93; cold-all-query ratio 15.28--35.97$\times$ & repair breadth \\
After-ingest repair & Qwen fixed \(K=1\) repair breadth & cold/repaired follow-up & 9.48--20.37$\times$ & choice 0/93; correct 0/93; pathological follow-ups 0/62 & repair breadth \\
After-ingest raw & Qwen raw warm reuse, 16f & cold/cached follow-up & 91.1$\times$ & 0.807\,s median; $\Delta$acc +0.000 & aggregate-clean; unrepaired \\
After-ingest raw & Qwen raw warm reuse, 8f & cold/cached follow-up & 47.2$\times$ & 0.815\,s median; $\Delta$acc -0.048 & within criterion \\
First-pass & Gemma VideoMME 8f holdout & first-query E2E & 1.113$\times$ & $\Delta$acc +0.000 & aggregate-clean \\
First-pass & Gemma sparse vision, 32f short & first-query E2E & 1.316$\times$ & choice agreement 100\%; $\Delta$acc +0.000; dense parse 0; sparse parse 0 & \PaperStack{clean}{timed-skip} \\
First-pass & Gemma MVBench 8f holdout & first-query E2E & 1.407$\times$ & $\Delta$acc -0.033; timing-attribution caveat & advisory \\
First-pass & Qwen sparse vision, 16f \(kr_V=0.85\) & first-query E2E & 1.032$\times$ & $\Delta$acc -0.050; vision reduction 13.6\%; parse failures 0; agreement 81.7\% & format gate; low gain \\
\bottomrule
\end{tabularx}
\end{table}

\noindent We present results in systems order: the high-leverage after-ingest
regime first, the first-pass sparse-vision regime and its share ceiling second,
and then the composition and paired-drift boundaries that prevent multiplying
headline numbers.

\subsection{C-PERSIST: After-Ingest Reuse Is The Big-Number Regime}

Persistent key/value (KV) attention cache reuse answers a different question:
once the first query on a video has already paid the full prompt cost, how cheap
can the next question become? Every result in this subsection is therefore an
after-ingest follow-up result: request-level savings depend on how many later
questions reuse the same video state, and the first query still pays full
freight. On Qwen2.5-VL-7B-Instruct-4bit, the answer is already large enough to matter.
At 8 frames, follow-up latency drops to 815\,ms median with 47.2$\times$
speedup and $-0.048$ aggregate accuracy delta. At 16 frames the speedup rises
to 91.1$\times$ while the accuracy delta returns to 0. Within the
$\leq$16f tested envelope, the regime is fast and empirically bounded: the
accepted region contains one clean point at 16f and one slightly worse but still
within tolerance at 8f. Past that point the curve keeps rising, but the fidelity
boundary appears.

\begin{figure}[H]
  \centering
  \includegraphics[width=0.97\linewidth]{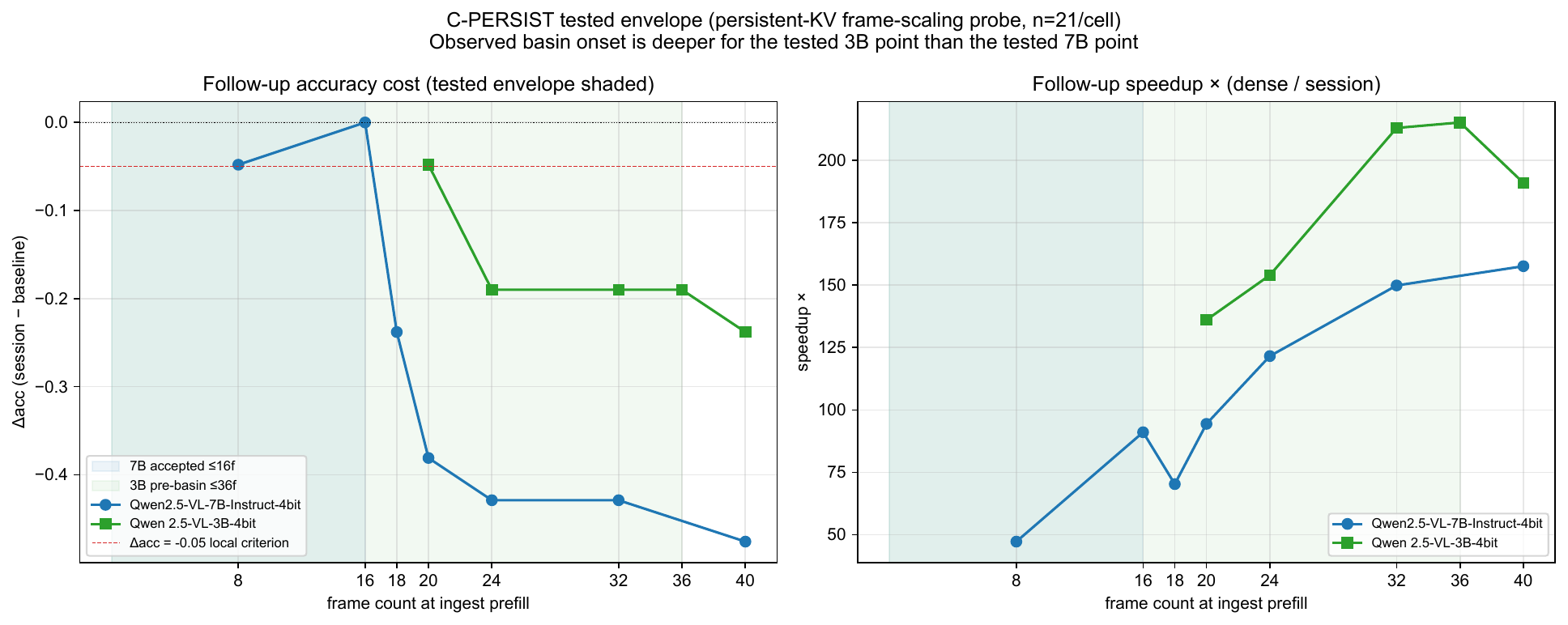}
\caption{Unrepaired persistent-KV frame-scaling envelope. The figure separates
  the raw warm-reuse probe from the repaired C-PERSIST breadth result in
  Table~\ref{tab:c-persist-repair}: 7B stays inside the tested raw envelope
  through 16 frames / 6.5k prefill tokens, while 3B trades paired fidelity for
  a wider pre-basin plateau through 36 frames / 14.5k prefill tokens.}
  \label{fig:c-persist-safe-budget}
\end{figure}

The 7B and 3B stories are therefore not interchangeable. 7B gives the clean
sub-second 16f point and a tested envelope through 16f. 3B does not. What 3B
gives instead is a wider pre-basin plateau: $\Delta$acc holds at $-0.19$
through 24f, 32f, and 36f before basin onset appears beyond 14.5k prefill
tokens. The useful mechanism is not an architecture-specific ``saturation
ceiling'' but a threshold story: onset depth moves with architecture, while
basin geometry remains the object to avoid or repair.

That repair path is local evidence, not just future work. Selective
re-prefill refreshes a small visual tail before a follow-up query. The simple
fixed policy is \(K=1\): refresh only the newest visual frame and reuse the
rest of the video prefix. The evidence is broad rather than anecdotal. Across 20f
short, medium, and long VideoMME slices plus a 32f short slice, fixed \(K=1\)
has no observed paired choice or correctness drift on 93 paired queries, no
observed pathological follow-up outputs on 62 repaired follow-ups, and
9.48--20.37$\times$ same-class follow-up speedup. The 32f point is important
because longer prefixes make the saved prefill work larger; \(K=1\) becomes
more useful as the prefix gets deeper, not less. The long-bucket point is also
important but weaker as an accuracy claim: it preserves a hard low-baseline
cell with 0/21 paired drift on the observed rows, so it is evidence for fidelity
repair under a difficult regime, not for high long-bucket model quality.
Paired drift is a reuse-preservation test, not a generosity discount for a weak
baseline: preserving the same hard examples under the paired-drift contract still constrains the cache
policy even when the frozen model's cold long-video accuracy is poor.

\begin{table}[H]
\centering
\caption{Selective re-prefill repair frontier. Fixed \(K=1\) is the no-coordination baseline; adaptive repaired-cache inheritance is the broad repair policy. Gains use cold/repaired follow-up or all-query denominators; setup-inclusive economics are in Table~\ref{tab:c-persist-setup-inclusive}.}
\label{tab:c-persist-repair}
\scriptsize
\renewcommand{\arraystretch}{\PaperTableStretch}
\begin{tabularx}{\linewidth}{@{}>{\raggedright\arraybackslash}p{0.18\linewidth} Y >{\raggedright\arraybackslash}p{0.22\linewidth} Y@{}}
\toprule
Policy & Scope & Follow-up median / cold gain & Validation signal \\
\midrule
Fixed \(K=1\) & 20f short/medium/long + 32f short & \shortstack[l]{6.78--10.14s\\9.48--20.37$\times$} & sessions with drift: 0/31; choice/correct 0/93 / 0/93; pathological follow-ups 0/62 \\
Adaptive repaired-cache inheritance & 20f short/medium/long + 32f short & \shortstack[l]{3.20--6.51s\\14.90--35.92$\times$ same-class\\15.28--35.97$\times$ all-query} & sessions with drift: 0/31; choice/correct 0/93 / 0/93; follow-up paired drift 0/62; second follow-up inherits repaired state; paired second-follow-up speedup 9.50$\times$ \\
\bottomrule
\end{tabularx}
\end{table}

The stronger repair result is adaptive. A failed variant that refreshed one
newest frame at the first follow-up and then omitted refresh at the second follow-up
fell back to the original unrepaired state and collapsed every
third query into the pathological basin: \(\Delta\)acc \(=-0.0952\), paired
correctness diffs 4/21, paired choice diffs 6/21, and pathological-like
outputs on 7/7 second follow-ups. The repaired adaptive variant changes only
the inheritance rule: refresh one newest frame at the first follow-up, then
let the second follow-up inherit the repaired cache state with no additional
visual re-prefill.

That rule has breadth. Across 20f short, medium, and long VideoMME slices
plus the 32f short slice, adaptive repair has no observed paired choice or
correctness drift on 93 paired queries, no observed pathological follow-up
outputs, and 14.90--35.92$\times$ same-class follow-up speedup
(15.28--35.97$\times$ cold-all-query / repaired-follow-up ratio). This is a
finite tested envelope, not a population proof: the query-level rule-of-three
bound is about 3.2\%, and the conservative session-level view is 0/31 sessions
with any paired drift. The 32f short cell is the largest local efficiency point
because the second follow-up inherits the repaired state instead of paying
another newest-frame re-prefill.

A timing attribution on the short-slice runs makes that concrete.
On the seven paired second follow-ups, fixed \(K=1\) has to re-prefill the
newest-frame tail again: median latency 6.65\,s, median tail prompt
451 tokens, and median prefix coverage 0.944. Adaptive repair reuses the
already-repaired cache and appends mostly the question text: median latency
0.675\,s, median tail prompt 50 tokens, and median prefix coverage 0.994. The
paired median second-follow-up fixed/adaptive speedup is therefore
9.50$\times$, with an 88.9\% median tail-token reduction. This explains why
adaptive dominates fixed \(K=1\) without changing the fidelity contract.

Sampler variation does not explain the result. On the seven-clip short-slice
mechanism cell, a five-temperature sweep and a \(3\times3\) seed/temperature
cross-product keep baseline quality in range, preserve the fidelity and format
gates, and retain large follow-up speedups. Two temperature-sweep cells still
show small paired diffs, so these are short-slice robustness checks, not part
of the 0/93 breadth claim.

\begin{table}[H]
\centering
\caption{Adaptive C-PERSIST sampler-temperature sweep on the short-slice mechanism cell.}
\label{tab:c-persist-sampler-stability}
\scriptsize
\renewcommand{\arraystretch}{\PaperTableStretch}
\begin{tabular}{@{}r r r r r r r@{}}
\toprule
\(T\) & Baseline & Session & \(\Delta\)acc & Choice diffs & Correct diffs & Speedup \\
\midrule
0.0 & 17/21 & 17/21 & +0.000 & 0 & 0 & 24.91$\times$ \\
0.5 & 17/21 & 17/21 & +0.000 & 0 & 0 & 25.39$\times$ \\
0.7 & 18/21 & 17/21 & -0.048 & 2 & 1 & 20.75$\times$ \\
1.0 & 18/21 & 17/21 & -0.048 & 1 & 1 & 23.17$\times$ \\
1.5 & 16/21 & 16/21 & +0.000 & 0 & 0 & 24.66$\times$ \\
\bottomrule
\end{tabular}
\end{table}

\begin{table}[H]
\centering
\caption{Adaptive C-PERSIST sampler-seed cross-product on the seven-clip, 21-pair short-slice cell. All nine seed/temperature cells pass the sampler robustness gate; small paired diffs remain bounded and are not part of the 0/93 breadth claim.}
\label{tab:c-persist-sampler-seed-sweep}
\small
\renewcommand{\arraystretch}{\PaperTableStretch}
\begin{tabular}{@{}r r r r r r@{}}
\toprule
\(T\) & Seed cells & Max \(|\Delta\)acc\(|\) & Max choice diffs & Max correctness diffs & Min baseline acc \\
\midrule
0.5 & 3 pass & 0.048 & 1/21 & 1/21 & 0.762 \\
1.0 & 3 pass & 0.048 & 1/21 & 1/21 & 0.714 \\
1.5 & 3 pass & 0.000 & 1/21 & 0/21 & 0.667 \\
\bottomrule
\end{tabular}
\end{table}

The follow-up envelope is no longer limited to two-turn sessions. A many-turn
stress repeats the same three VideoMME short questions across 10-, 20-, and
50-turn schedules on seven 20-frame videos. It tests cache horizon, not natural
dialogue: the dense baseline rows are deterministic replicas of 21 unique
stateless dense runs, which makes them valid for turn-matched drift pairing but
not independent timing samples. Adaptive repair and scheduled refresh show
0/343 paired choice drift and 0/343 paired correctness drift through the
50-turn horizon, with median follow-up latency below 0.8\,s. Fixed \(K=1\)
stays below the 3\% gate but shows sparse nonzero drift, 3/343 choice diffs and
2/343 correctness diffs. The result removes the obvious two-follow-up cliff;
true conversational stability remains future work.

\begin{table}[H]
\centering
\caption{Many-turn C-PERSIST stress on seven 20f short VideoMME videos. The stress cycles the same three questions through a 50-turn stateless question schedule, so it tests cache-horizon stability under repeated follow-ups rather than natural dialogue. The 560 dense baseline rows are deterministic replicas of 21 unique stateless dense runs for turn-matched pairing, not independent timing samples.}
\label{tab:c-persist-many-turn}
\scriptsize
\renewcommand{\arraystretch}{\PaperTableStretch}
\begin{tabularx}{\linewidth}{@{}>{\raggedright\arraybackslash}p{0.18\linewidth} r r >{\raggedright\arraybackslash}p{0.135\linewidth} r Y@{}}
\toprule
Policy & Horizon & Follow-ups & \PaperStackHead{Choice drift}{Correct drift} & Median follow-up & Gate / post-repair check \\
\midrule
fixed \(K=1\) & 50 & 343 & \PaperStack{3/343}{2/343} & 7.362\,s & passes 3\% gate but has nonzero repeated-question drift; no pathologies \\
adaptive repair & 50 & 343 & \PaperStack{0/343}{0/343} & 0.771\,s & post-repair choice 0/336; correct 0/336; no cliff \\
scheduled refresh-10 & 50 & 343 & \PaperStack{0/343}{0/343} & 0.706\,s & post-repair choice 0/308; correct 0/308; no cliff \\
\bottomrule
\end{tabularx}
\end{table}

The stricter prompt-variation stress deliberately breaks the repeated-question
assumption. Each follow-up prompt includes the previous canonical dense answer,
and the dense and cached arms receive identical prompt hashes. Conservative
fixed \(K=1\) remains paired-drift clean through the 20-turn horizon:
0/133 follow-up choice drift and 0/133 correctness drift. The more aggressive
adaptive and refresh-10 policies each show bounded but real drift, 6/133
choice/correctness diffs (4.51\%), while keeping median follow-up latency near
0.7\,s. Drift is split across early and late follow-ups rather than
concentrated at the end, so the result is a measured operating envelope: no
observed late-turn cliff through horizon 20, but no claim of universal dialogue
stability for the aggressive policies.

\begin{table}[H]
\centering
\caption{Dense-answer-anchored C-PERSIST stress. Each follow-up prompt includes the previous canonical dense answer and the dense/cached arms receive identical prompt hashes. This is a content-conditioned prompt-variation stress, not natural dialogue. Parenthesized speedups in the median-follow-up column use an approximate 80\,s cold-dense reference for readability.}
\label{tab:c-persist-dense-anchored}
\scriptsize
\renewcommand{\arraystretch}{\PaperTableStretch}
\begin{tabularx}{\linewidth}{@{}>{\raggedright\arraybackslash}p{0.14\linewidth} r r >{\raggedright\arraybackslash}p{0.155\linewidth} >{\raggedright\arraybackslash}p{0.18\linewidth} Y@{}}
\toprule
Policy & Horizon & Follow-ups & \PaperStackHead{Choice drift}{Correct drift} & \PaperStackHead{Median follow-up}{speedup} & Interpretation \\
\midrule
fixed \(K=1\) & 20 & 133 & \PaperStack{0/133}{0/133 (0.00\%)} & 5.329\,s ($\sim$15.0$\times$) & passes 3\% gate; no observed prompt-variation drift \\
adaptive repair & 20 & 133 & \PaperStack{6/133}{6/133 (4.51\%)} & 0.698\,s ($\sim$114.6$\times$) & fails 3\% gate; early/late choice drift 3/63 and 3/70; no cliff \\
scheduled refresh-10 & 20 & 133 & \PaperStack{6/133}{6/133 (4.51\%)} & 0.709\,s ($\sim$112.8$\times$) & fails 3\% gate; early/late choice drift 3/63 and 3/70; no cliff \\
\bottomrule
\end{tabularx}
\end{table}

Warm-only multipliers are the right follow-up denominator, but they are not the
whole serving economics. Table~\ref{tab:c-persist-setup-inclusive} folds the
cold first query back into the session. On the clean 7B/16f raw reuse point, a
single-query session is essentially neutral because the first query still pays
full freight; the same row reaches 2.05$\times$ at two total queries,
9.33$\times$ at ten total same-video queries, and 32.17$\times$ at fifty. This
is the product-facing calculus: after-ingest reuse is valuable when sessions
have follow-up questions, not because the first question became cheap.

\begin{table}[H]
\centering
\caption{Setup-inclusive same-video follow-up economics. \emph{Warm} is mean per-follow-up speedup; \(Q=k\) is session speedup for \(k\) same-video queries. Qwen counts one measured cold first query plus \(Q-1\) C-PERSIST follow-ups; Gemma counts prefix warm-up plus \(Q\) follow-ups for denominator comparison. \(\Delta\)acc is paired session-vs-baseline accuracy delta where available; Gemma zeros reflect zero paired correctness diffs.}
\label{tab:c-persist-setup-inclusive}
\small
\renewcommand{\arraystretch}{\PaperTableStretch}
\begin{tabular}{@{}l r r  r r r r r@{}}
\toprule
Cell & $\Delta$acc & Warm & Q=1 & Q=2 & Q=5 & Q=10 & Q=50 \\
\midrule
\multicolumn{8}{@{}l}{\emph{Local Qwen C-PERSIST: cold first query + follow-ups}} \\
7B / 16f & +0.00 & 83.00$\times$ & 1.04 & 2.05 & 4.94 & 9.33 & 32.17 \\
7B / 18f & -0.24 & 45.50$\times$ & 0.96 & 1.88 & 4.43 & 8.07 & 23.60 \\
7B / 20f & -0.38 & 39.58$\times$ & 1.06 & 2.06 & 4.77 & 8.52 & 22.89 \\
7B / 24f & -0.43 & 124.12$\times$ & 1.08 & 2.14 & 5.22 & 10.01 & 37.85 \\
7B / 32f & -0.43 & 153.81$\times$ & 0.99 & 1.96 & 4.80 & 9.32 & 37.50 \\
3B / 20f & -0.05 & 132.21$\times$ & 1.05 & 2.08 & 5.09 & 9.79 & 37.78 \\
\midrule
\multicolumn{8}{@{}l}{\emph{Gemma 26B prefix snapshots: after-warm prefix snapshot + follow-ups}} \\
Gemma 26B / 8f & +0.00 & 6.81$\times$ & 0.85 & 1.52 & 2.84 & 4.01 & 5.97 \\
Gemma 26B / 32f & +0.00 & 18.19$\times$ & 1.01 & 1.92 & 4.14 & 6.74 & 13.58 \\
\bottomrule
\end{tabular}
\end{table}

\begin{figure}[H]
  \centering
  \IfFileExists{generated/figures/c_persist_timeline.pdf}{
    \includegraphics[width=0.98\linewidth]{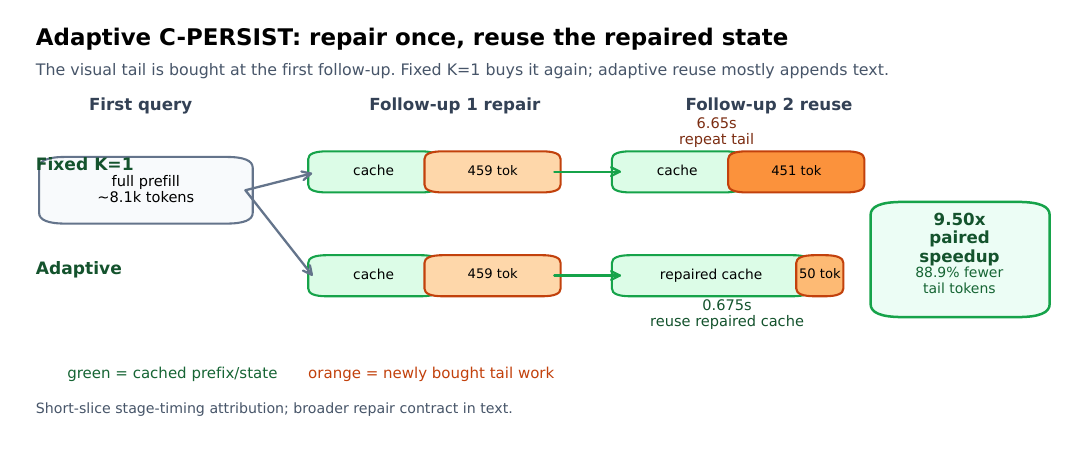}
  }{
    \fbox{\parbox{0.9\linewidth}{Figure unavailable in this source snapshot.}}
  }
  \caption{Adaptive C-PERSIST timing attribution. Fixed \(K=1\) buys the
  newest-frame tail again at the second follow-up; adaptive repair mostly
  appends text from the repaired cache. The visual shows the short-slice
  timing mechanism behind the 9.50\(\times\) paired second-follow-up speedup;
  the broader 0/93 fidelity result is in Table~\ref{tab:c-persist-repair}.}
  \label{fig:c-persist-timeline}
\end{figure}

\subsection{C-VISION: First-Pass Gains Are Real, But Share-Limited}

Fresh-video first-pass gains are smaller because they are share limited. Gemma
vision-tower pruning still produces measured end-to-end gains before any video
cache enters the picture. VideoMME 8f is the clean holdout anchor:
1.113$\times$ first-query speedup with zero aggregate accuracy delta. MVBench
8f has the larger observed multiplier at 1.407$\times$, but it remains
advisory because decode-stage imbalance and generation-time residuals prevent
clean attribution. Normalizing only the decode imbalance would lower the
observed multiplier only to about 1.40$\times$, so the row remains useful but
high residual. TOMATO 8f is advisory as well: the holdout rerun reports
1.194$\times$, but decode-stage imbalance misses the 100\,ms advisory gate by
19.7\,ms.

The denominator, not the absolute multiplier, is the portable claim. The same
pruning mechanism gives smaller gains on VideoMME than on MVBench because
VideoMME is less vision-dominated at this geometry. A matched Qwen VideoMME 8f
cross-architecture probe at the same \(L=2\), \(kr_V=0.50\) operating point
lands at 1.044$\times$ observed versus
1.043$\times$ predicted, with aggregate accuracy delta $-0.033$ on \(n=30\)
dev. The absolute lift is smaller because Qwen's dense
\(V_{\mathrm{share}}\) at this geometry is only about 10\%. What transfers is
the share-weighted accounting rule, not a benchmark-independent multiplier.

The new measured sparse-vision runs turn that accounting rule into a sharper
systems result. They do not just replay dense features; they skip timed
vision-tower work. On Gemma, the 8f/16f/32f sweep has zero paired accuracy
delta and 100\% choice agreement at all three frame budgets. The full sweep is
not an all-gates headline because dense and sparse arms share parse failures,
and the 8f row misses the ceiling tolerance by +0.062$\times$. The clean
operating point is narrower and more interesting: the 32f short bucket reaches
1.316$\times$ first-query speedup with no paired drift on 20 items, no parse
failures, 42.2\% vision-time reduction, and a \(-0.012\)\(\times\) ceiling
gap.

Qwen exposes the architecture/configuration boundary. Aggressive 16f pruning
at \(kr_V=0.50\) collapses format and aggregate accuracy. Raising the keep rate
recovers the model monotonically; \(kr_V=0.85\) is inside the tested aggregate
accuracy and format gates, not a paired-identity gate, and is ceiling-consistent.
The price is that vision-time reduction falls to
13.6\% and the observed end-to-end gain is only 1.032$\times$. The right
paper claim is therefore an envelope: Gemma supplies the strongest measured
sparse-execution cell, while Qwen shows that the same timed-skip idea is
configuration-sensitive and can trade fidelity for too little saved vision
work.

A trivial random-keep sanity baseline confirms that the Qwen scorer is not
arbitrary. At the same 8f, \(L=2\), \(kr_V=0.50\) keep-rate on VideoMME dev30,
the structured magnitude scorer reaches 0.500 accuracy versus a four-seed
uniform-random mean of 0.358 (range 0.333--0.367), with dense at 0.533. That is
a +14.2\,pp structured-vs-random gap at matched keep-rate, not a peer-method
comparison and not a matched-runtime claim.

\begin{table}[H]
\centering
\caption{Random-keep sanity check at matched keep-rate. Same Qwen2.5-VL-7B-Instruct-4bit VideoMME dev30 setup, frame count, and vision-tower cut layer \(L=2\). The \emph{magnitude\_norm} row is the structured Qwen scorer; the \emph{uniform\_random} row averages trivial-baseline seeds at the same keep-rate. The \(\Delta\)acc column is relative to dense. The random row averages 4 seeds. The structured-vs-random gap is +14.2\,pp over the four random seeds evaluated. Wall-clock columns are measured but not matched-runtime peer-method claims.}
\label{tab:competitor-positioning}
\small
\renewcommand{\arraystretch}{\PaperTableStretch}
\begin{tabular}{@{}l c c r r r r@{}}
\toprule
Arm & \(kr_V\) & N & Acc & $\Delta$acc vs dense & Vision ms & E2E ms \\
\midrule
Unpatched (dense reference) & 1.00 & 30 & 0.533 & +0.0\,pp & 8790 & 85437 \\
magnitude\_norm (structured scorer) & 0.50 & 30 & 0.500 & -3.3\,pp & 5294 & 81862 \\
uniform\_random mean (4 seeds) & 0.50 & 30$\times$4 & 0.358 & -17.5\,pp & 4366 & 76581 \\
\bottomrule
\end{tabular}
\end{table}

\begin{figure}[H]
  \centering
  \includegraphics[width=0.97\linewidth]{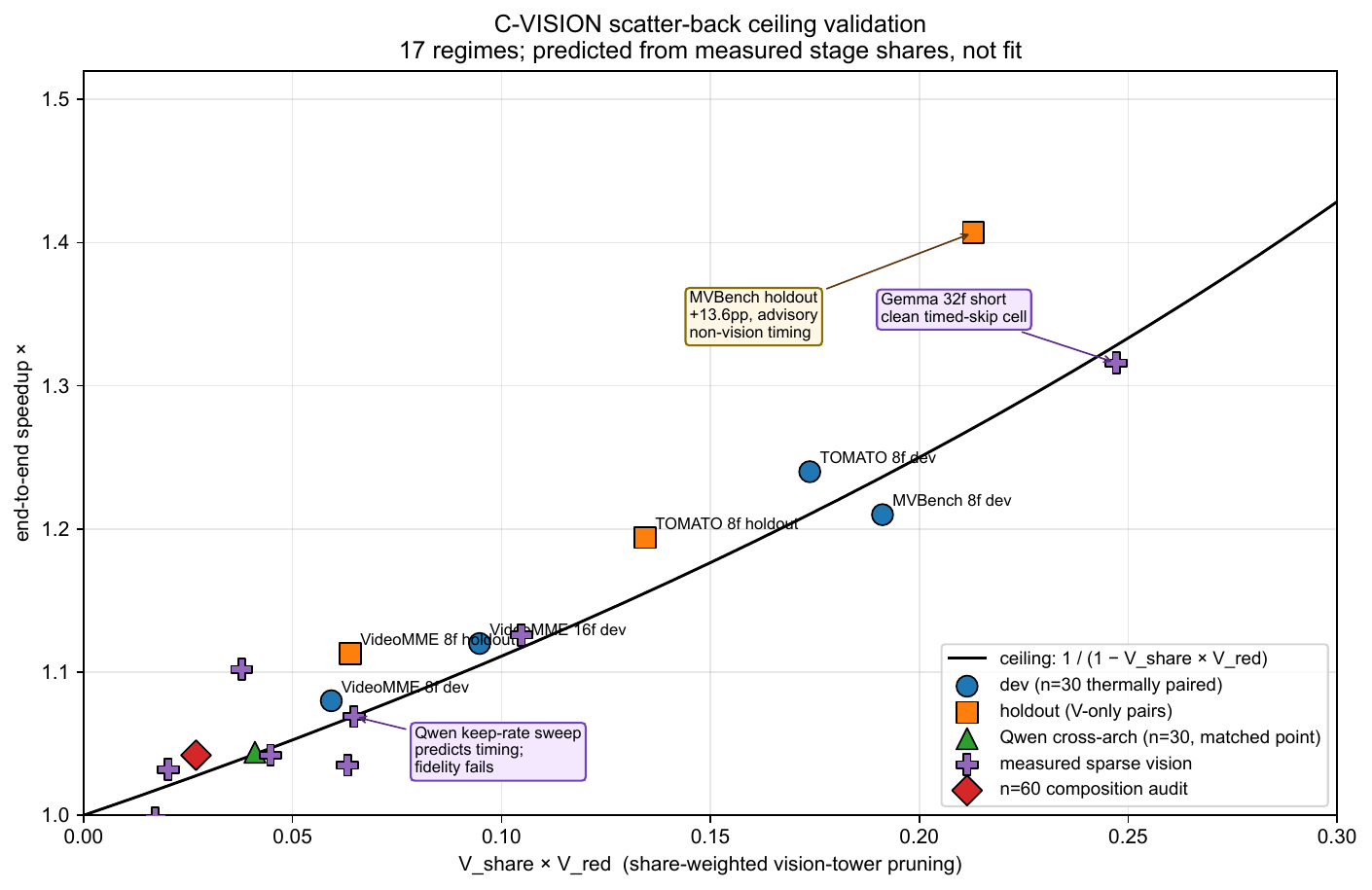}
  \caption{C-CEILING share-model validation. Predicted speedup uses dense
  vision share and pruned-run vision reduction, then compares with independently
  observed end-to-end speedup. The n=60 composition audit and low-share Qwen
  point show denominator binding; measured-sparse markers add Qwen
  timing-validating fidelity failures plus the low-gain 16f boundary. The clean
  measured sparse-execution headline is Gemma 32f short in
  Table~\ref{tab:headline-results}.}
  \label{fig:vshare-ceiling}
\end{figure}

\begin{table}[H]
\centering
\caption{Measured sparse vision execution on Gemma. The timed path skips real vision-tower work. Full-sweep rows preserve paired accuracy and agreement but share parse failures; only 32f short is parse-clean and supplies the 1.316\(\times\) headline cell.}
\label{tab:gemma-measured-sparse-vision}
\scriptsize
\renewcommand{\arraystretch}{\PaperTableStretch}
\begin{tabularx}{\linewidth}{@{}l r r r >{\centering\arraybackslash}p{0.095\linewidth} r r r Y@{}}
\toprule
Frames / slice & \(n\) & \(\Delta\)acc & Agree & \PaperStackHead{Parse}{fails} & \(V_{\mathrm{red}}\) & Observed & Residual & Ceiling / format verdict \\
\midrule
8f & 60 & +0.000 & 100\% & \PaperStack{dense 11}{sparse 11} & 48.2\% & 1.102$\times$ & +0.062 & ceiling miss; format not clean (matched parse failures) \\
16f & 60 & +0.000 & 100\% & \PaperStack{dense 3}{sparse 3} & 40.6\% & 1.035$\times$ & -0.032 & ceiling pass; format not clean (matched parse failures) \\
32f & 60 & +0.000 & 100\% & \PaperStack{dense 4}{sparse 4} & 43.3\% & 1.126$\times$ & +0.009 & ceiling pass; format not clean (matched parse failures) \\
32f short & 20 & +0.000 & 100\% & \PaperStack{dense 0}{sparse 0} & 42.2\% & 1.316$\times$ & -0.012 & clean sparse-execution operating point \\
\bottomrule
\end{tabularx}
\end{table}

\begin{table}[H]
\centering
\caption{Measured sparse vision execution on Qwen. The 8f sweep validates C-CEILING timing but has no fidelity-preserving, speed-positive point. The 16f sweep recovers to within the aggregate-accuracy/format gates at \(kr_V=0.85\), but not paired identity and not the vision-reduction gate.}
\label{tab:qwen-measured-sparse-vision}
\scriptsize
\renewcommand{\arraystretch}{\PaperTableStretch}
\begin{tabularx}{\linewidth}{@{}l r r r r r r r Y@{}}
\toprule
Setting & \(n\) & \(\Delta\)acc & Agree & Parse fails & \(V_{\mathrm{red}}\) & Observed & Residual & Ceiling / quality verdict \\
\midrule
8f, $kr_V$=0.25 & 60 & -0.217 & 58.3\% & 2 & 64.8\% & 1.069$\times$ & +0.000 & timing-validating fidelity/format fail \\
8f, $kr_V$=0.50 & 60 & -0.067 & 71.7\% & 0 & 44.8\% & 1.042$\times$ & -0.005 & timing-validating fidelity-contract fail \\
8f, $kr_V$=0.75 & 60 & -0.083 & 83.3\% & 0 & 17.2\% & 0.998$\times$ & -0.020 & timing-validating fidelity-contract fail; no E2E gain \\
16f, $kr_V$=0.65 & 60 & -0.283 & 46.7\% & 11 & 36.1\% & 1.093$\times$ & +0.036 & recovery sweep \\
16f, $kr_V$=0.75 & 60 & -0.100 & 75.0\% & 0 & 27.0\% & 1.065$\times$ & +0.023 & recovery sweep \\
16f, $kr_V$=0.85 & 60 & -0.050 & 81.7\% & 0 & 13.6\% & 1.032$\times$ & +0.011 & within aggregate-accuracy/format gates; low-gain boundary \\
\bottomrule
\end{tabularx}
\end{table}

\begin{table}[H]
\centering
\caption{Numeric residuals for the original C-CEILING cells. The prediction is
computed from measured share terms, not fit to the observed end-to-end
multiplier. The newer measured sparse-execution rows are plotted in
Figure~\ref{fig:vshare-ceiling} and tabulated in
Tables~\ref{tab:gemma-measured-sparse-vision}
and~\ref{tab:qwen-measured-sparse-vision}. Residual is observed minus
predicted speedup. MVBench's advisory residual is not cleanly attributable, so
it is not used as a clean first-pass headline.}
\label{tab:vshare-ceiling-residuals}
\scriptsize
\renewcommand{\arraystretch}{\PaperTableStretch}
\begin{tabularx}{\linewidth}{@{}Y r r r r r Y@{}}
\toprule
Cell & \(V_{\mathrm{share}}\) & \(V_{\mathrm{red}}\) & Pred. & Obs. & Resid. & Evidence \\
\midrule
VideoMME 8f dev & 0.152 & 0.390 & 1.062 & 1.080 & +1.8pp & Gemma dev \\
VideoMME 16f dev & 0.243 & 0.390 & 1.105 & 1.120 & +1.5pp & Gemma dev \\
MVBench 8f dev & 0.478 & 0.400 & 1.237 & 1.210 & -2.7pp & Gemma dev \\
TOMATO 8f dev & 0.407 & 0.427 & 1.214 & 1.240 & +2.6pp & Gemma dev \\
VideoMME 8f holdout & 0.154 & 0.413 & 1.068 & 1.113 & +4.5pp & aggregate-clean \\
MVBench 8f holdout & 0.452 & 0.471 & 1.271 & 1.407 & +13.6pp & advisory \\
TOMATO 8f holdout & 0.384 & 0.350 & 1.155 & 1.194 & +3.9pp & advisory \\
Qwen VideoMME 8f dev & 0.103 & 0.398 & 1.043 & 1.044 & +0.1pp & matched cross-arch \\
Qwen measured sparse vision 8f \(kr_V=0.50\) & 0.100 & 0.448 & 1.047 & 1.042 & -0.5pp & boundary; fidelity fail \\
VideoMME n=60 composition audit & 0.063 & 0.430 & 1.028 & 1.042 & +1.4pp & composition null \\
\bottomrule
\end{tabularx}
\end{table}

Across the ceiling table and the newer sparse-execution tables, the share model stays predictive enough to
be useful and honest, but it is not a literal no-residual theorem. Advisory
cells can sit above or below the prediction when non-vision timing varies
across paired arms. The holdout spread is wider than
the first dev tranche made it look, and the matched Qwen point sits low-left
where its smaller \(V_{\mathrm{share}}\) says it should. Gemma's 32f short cell
is the cleanest measured sparse-execution operating point; Qwen's 8f keep-rate
sweep is a timing-validation/fidelity-failure boundary, and its 16f \(kr_V=0.85\)
point is the low-gain aggregate-accuracy/format boundary. That is why the paper describes
benchmark-, architecture-, and protocol-sensitive magnitude rather than
benchmark-invariant first-pass gain. \(V_{\mathrm{share}}\) governs magnitude,
while \(V_{\mathrm{red}}\) itself is benchmark- and configuration-conditional.
In the original holdout and cross-architecture C-CEILING cells,
\(V_{\mathrm{red}}\) spans 0.350--0.471; the newer measured sparse-execution
rows broaden that configuration range and are reported separately in
Tables~\ref{tab:gemma-measured-sparse-vision}
and~\ref{tab:qwen-measured-sparse-vision}. That softening is not a retreat. It
is what the holdouts, measured sparse-execution rows, and cross-architecture
probe earned.

\subsection{What Stacks and What Does Not}

The measured first-pass gains and the after-ingest gains complement rather than
compete with each other. First-pass pruning says there is real work to skip on
fresh videos, but the ceiling is share-limited. Persistent-KV says that once a
video has already been ingested, the next question is a much larger reuse
opportunity. The routing and bridge results explain why they do not simply
multiply: temporal placement of fresh compute decides whether the answer
survives, and the reused cache state can matter even when the visible first
answer looks fine.

The composition bridge makes the same warning concrete. On Qwen2.5-VL-7B-Instruct-4bit
at 8 frames, the original session/streaming stack that combines \(L=2\),
\(kr_V=0.50\) vision pruning with persistent-KV reuse over a VideoMME
dev+holdout union (\(n=57\) sessions, 171 queries) was a hard negative. The
dense-first-query reference improves the failure mode but remains a near-miss
rather than a deployable policy: cold aggregate accuracy is 0.561 and the streaming path
is 0.503, a \(-0.0585\) delta with 95\% paired CI roughly \([-0.117, 0.000]\),
at 2.79$\times$ paired amortized end-to-end speedup. The later instrumented
cache-reuse rerun preserves the faster 3.02$\times$ speed profile while showing
that follow-up vision pruning is inactive under prompt-cache reuse. The
important repair is that aggregate first-query correctness returns to parity,
with zero parse failures and zero degenerate outputs. The remaining loss is
follow-up state damage.

The long-bucket sweep closes the tested admission family. None of the tested
Q0 keep rates \(\{0.67,0.75,0.80,0.85,0.90\}\) reaches the target
fidelity/speed band. More importantly, the instrumentation shows that the follow-up
``pruning'' arm is mechanically inactive under prompt-cache reuse:
\texttt{vision\_pruning\_active\_fraction} is 0.0 and all follow-up image
tokens are cache-served. The result is therefore Q0 admission plus cache-state
reuse, not active follow-up vision pruning.

\begin{table}[H]
\centering
\caption{Qwen composition boundary. The tested admission family has no deployable composition point: cache reuse and invalidation hit the same aggregate loss through different drift sets, and only cache reuse preserves the three-query speed profile. Speedup is paired three-query E2E versus cold all-query execution; follow-up vision active is the measured activity fraction; \(kr_{Q0}\) is first-query visual-token keep rate.}
\label{tab:qwen-bridge-boundary}
\small
\renewcommand{\arraystretch}{\PaperTableStretch}
\par\centerline{\resizebox{0.95\linewidth}{!}{%
\begin{tabular}{@{}l r r r r r r r r@{}}
\toprule
Policy & Sessions & Queries & \(\Delta\)acc & Q0 \(\Delta\) & Follow-up \(\Delta\) & 3-query E2E & Follow-up vision active & Degenerate outputs \\
\midrule
cache reuse & 57 & 171 & -0.058 & +0.000 & -0.088 & 3.02$\times$ & 0.0 & 0 \\
cache invalidated & 57 & 171 & -0.058 & +0.000 & -0.088 & 1.06$\times$ & 1.0 & 0 \\
dense Q0 reference & 57 & 171 & -0.058 & +0.000 & -0.088 & 2.79$\times$ & -- & 0 \\
long \(kr_{Q0}=0.67\) & 18 & 54 & -0.130 & -0.111 & -0.139 & 3.12$\times$ & 0.0 & 0 \\
long \(kr_{Q0}=0.75\) & 18 & 54 & -0.185 & -0.111 & -0.222 & 3.37$\times$ & 0.0 & 1 \\
long \(kr_{Q0}=0.80\) & 18 & 54 & -0.185 & -0.111 & -0.222 & 3.38$\times$ & 0.0 & 0 \\
long \(kr_{Q0}=0.85\) & 18 & 54 & -0.130 & +0.000 & -0.194 & 3.35$\times$ & 0.0 & 0 \\
long \(kr_{Q0}=0.90\) & 18 & 54 & -0.130 & +0.000 & -0.194 & 3.34$\times$ & 0.0 & 0 \\
\bottomrule
\end{tabular}%
}}
\end{table}

The mechanism is sharper than a generic failure. At high Q0 keep rates
(\(kr_{Q0}=0.85\) and 0.90), aggregate Q0 accuracy parity returns, but
follow-up accuracy still drops by 19.4 points. The completed dense-Q0 reference,
cache-reuse, and cache-invalidated tests make the boundary cleaner. Cache reuse
preserves the 3.02$\times$ three-query speed profile but leaves the same net
\(-0.0585\) aggregate loss. Cache invalidation forces follow-up V-pruning to
fire on every follow-up, collapses the speedup to 1.06$\times$, and lands on
the same net aggregate loss. The any-paired-drift sets are not identical
(Jaccard overlap 0.3125; intersection over the union of examples with either
choice drift or correctness drift). The claim is therefore not identical
row-level failure. It is stronger and narrower: two different mechanisms reach
the same aggregate boundary, while only the cache-reuse path retains the
wall-clock profile.

A cache-distance audit supports the mechanism diagnosis: no-prune reuse stays
dense-like, while the pruned/recomputed path materially diverges in later KV
state, especially values. The probe does not give a deployable row-level drift
predictor, but it explains why aggregate parity is too weak a composition
contract.

A low-FPS dense baseline is the corresponding ``why not just sample fewer
frames?'' control. On the same 57-session / 171-query VideoMME union, 4f dense
is 1.21$\times$ faster than the 8f dense reference, but the overall accuracy
delta is ambiguous rather than clean: \(-0.0526\), 95\% paired CI
\([-0.1228,0.0175]\), with no parse failures and 72.5\% choice agreement. The
short bucket is the hard failure (\(-0.193\), CI \([-0.316,-0.070]\)), while
medium and long are inconclusive or neutral. This baseline matters because it
is not a strawman. Some workloads can buy speed by lowering frame count; the
paper's stronger claim is that low-FPS dense is an accuracy-risky baseline that
does not replace cache repair or measured sparse execution.

The richer drift-prediction scout is also negative in the useful direction.
Dense answer margin has signal but not enough coverage to be a deployable
guard, and adding entropy, duration, turn index, prompt length, and policy
source does not close the gap on the held-out split. The nominal label-free
feature is empirically answer-aware: its top-second-gap feature is identical
to the dense answer margin. We therefore do not claim a learned cache-validity
guard. The result says the next guard needs genuinely independent state
features, not just denser logprob plumbing.

The key compositional result is negative and useful: the n=60
stacked-regime experiment is a ceiling-matched null, not a
super-ceiling win. The observed 1.042$\times$ composition-audit result lands where the
measured fixed-fraction terms place it. That closes the tempting
``multiply-the-headlines'' interpretation. C-VISION and C-PERSIST matter, but
they matter on different queries and under different ceilings.

\subsection{Aggregate Accuracy Is Not Enough}

Several of the most informative results in the paper would be invisible if we
reported only aggregate accuracy. In the composition root-cause decomposition,
V-only Q0 pruning changes 6/10 choices and 6/10 correctness labels on the
short scout even though the broader C-VISION operating point can look
aggregate-neutral elsewhere. In the Gemma second-architecture routing check,
MVBench ties aggregate accuracy (dense and cached both 6/30) but
changes 8/30 answer choices across four motion subgroups. In the unrepaired
20f persistent-KV basin, 11/21 choices and 10/21 correctness labels change,
with 13/21 pathological-like outputs concentrated on follow-ups.

\begin{table}[H]
\centering
\caption{Paired-drift examples. Aggregate accuracy can preserve a point estimate while answer identity, correctness on individual examples, or output format changes.}
\label{tab:paired-drift}
\small
\renewcommand{\arraystretch}{\PaperTableStretch}
\begin{tabularx}{\linewidth}{@{}Y r r r Y@{}}
\toprule
Source & \(N\) & Choice drift & Correctness drift & Mechanism signal \\
\midrule
Qwen first-query admission scout & 10 & 6/10 & 6/10 & first-query admission damage \\
Gemma MVBench holdout & 30 & 8/30 & 4/30 & aggregate tie hides answer drift; dense/cached acc. 0.200/0.200 \\
unrepaired persistent-KV 20f & 21 & 11/21 & 10/21 & follow-up cache basin; pathological-like 13/21 \\
\bottomrule
\end{tabularx}
\end{table}

That is the explanatory link across the paper's failures. C-CEILING tells us
whether a speedup can survive the denominator. Paired drift tells us whether
the answer survived for the same reasons. A method that only preserves aggregate
accuracy may still be uncertified for reuse, and a method that fails can fail by
ordinary answer drift, by first-query admission damage, or by a cache-state
basin. Those are different repairs.

\section{Routing Mechanism and Quality--Compute Frontier}
\label{sec:routing-mechanism}

The Qwen routing lane is the cleanest explanation for why training-free
anti-recomputation can work at all. It tells us what kind of temporal signal
is useful, what kind is too blunt, and where the semantic-substitution story
stops.

\subsection{Held-out Routing Frontier}

On TOMATO motion, the base policy matches dense-8 accuracy while spending
only 3.55 effective fresh frames instead of 8.0. On MVBench motion, the same
base policy reaches a higher point estimate than dense-6 while spending
less visual refresh: 0.600 accuracy at 4.06 effective fresh frames versus
0.567 at 6.0 fresh frames. At \(n=30\), that is a pointwise Pareto frontier
observation, not a powered benchmark win. Those are not wall-clock claims.
They are statements about preserved answer quality at lower effective visual
budget. The MVBench statement is against the dense-6 point at a
matched-or-lower fresh-frame budget; dense-8 remains the higher
absolute-accuracy point on that slice.

\subsection{Frontier Snapshot}

\begin{table}[H]
\centering
\caption{Qwen base-policy holdout frontier. Fresh is effective fresh frames; the rows compare answer quality at lower visual budget, not timed skipped work.}
\label{tab:lane-a-holdout}
\small
\renewcommand{\arraystretch}{\PaperTableStretch}
\begin{tabular}{lllrrrl}
\toprule
Benchmark & Policy & Relation & Acc. & Fresh & Comparator & Comparator acc. \\
\midrule
TOMATO & planner base & tie & 0.333 & 3.55 & dense-8 (8) & 0.333 \\
MVBench & planner base & frontier & 0.600 & 4.06 & dense-6 (6) & 0.567 \\
\bottomrule
\end{tabular}
\end{table}

\begin{figure}[H]
  \centering
  \IfFileExists{generated/figures/lane_a_pareto.pdf}{
    \includegraphics[width=\linewidth]{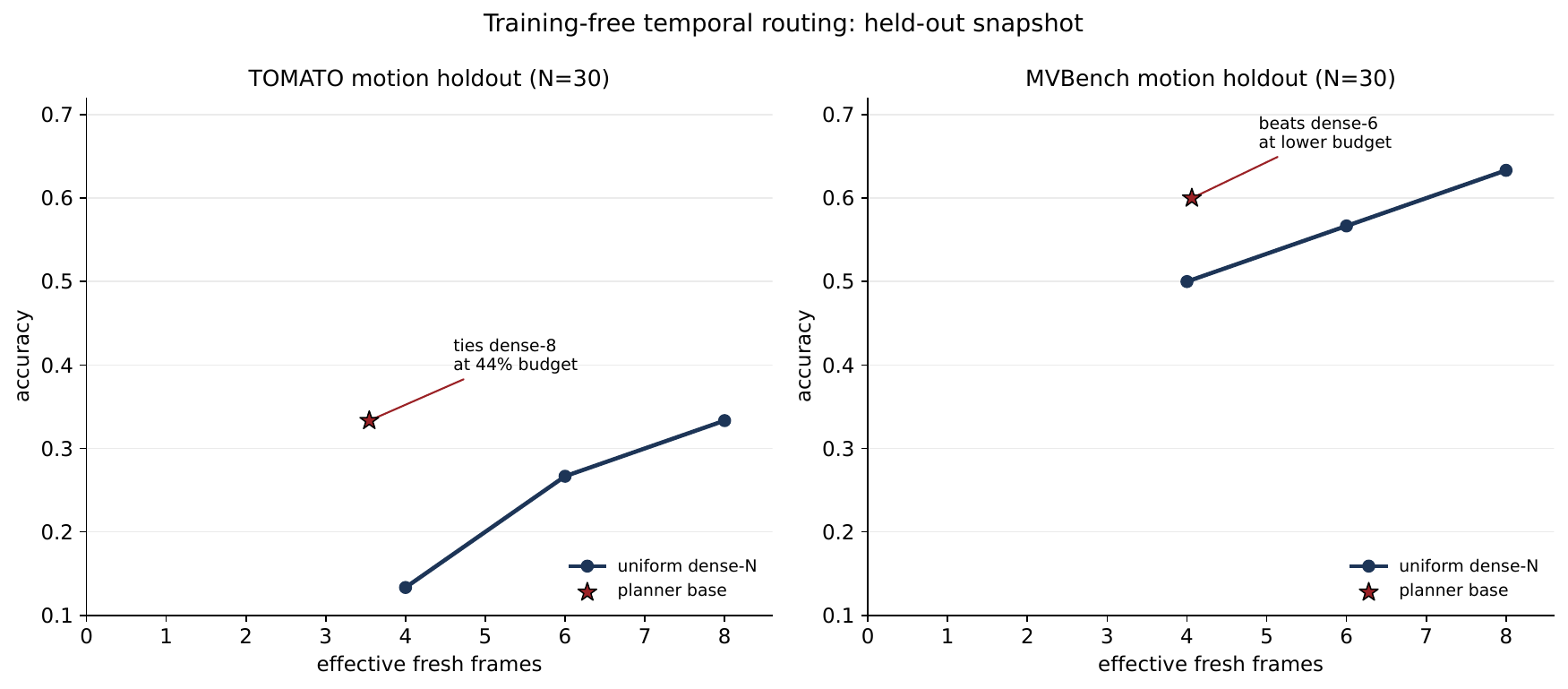}
  }{
    \fbox{\parbox{0.9\linewidth}{Figure unavailable in this source snapshot.}}
  }
\caption{Qwen base-policy routing frontier under dense-backend substitution.}
  \label{fig:lane-a-pareto}
\end{figure}

The takeaway is narrow by design: temporal placement can preserve answer
quality while buying fewer fresh frames on these motion holdouts. This is
mechanism evidence for anti-recomputation, not a deployable speedup claim and
not a universal routing policy. The timed skipped-work claims remain in
C-VISION, and the large latency claims remain in C-PERSIST.

\subsection{A Second Architecture Shows The Boundary}

The same planner is not governed by a single universal attention topology
where one pattern is always reliable and another is simply unusable. A Gemma
4-E4B-4bit holdout pair makes the boundary more useful. On TOMATO motion,
Gemma preserves aggregate accuracy
and passes the preregistered strict-agreement gate: dense and cached accuracy
are both 8/30, with 28/30 answer agreement. On MVBench motion, aggregate
accuracy still ties (both dense and cached are 6/30), but
strict agreement falls to 22/30 and misses the 0.85 gate. The eight MVBench
mismatches spread across action localization, fine-grained action, object
interaction, and moving direction rather than one brittle subgroup.

That split is part of the mechanism story. Qwen remains the high-stability
reference family for the tested routing cells. Gemma shows that feature
similarity and aggregate accuracy are not enough: answer-level reuse fidelity
is architecture-conditioned \emph{and} benchmark-conditioned. This is why the
paper reports paired answer drift instead of treating equal aggregate accuracy
as proof that reuse is semantically identical.

\subsection{Mechanism Diagnostics}

Two local diagnostics sharpen the mechanism story but are not main-claim
evidence here. The novelty-ranked dense baseline tests
whether the cached planner is merely selecting visually novel frames; the
feature-change oracle correlates blockwise pixel statistics with blockwise ViT
feature change on cached dense features. These diagnostics inform the
mechanism interpretation, but the reported claim rests on the holdout
frontier results.

The reported claim does not depend on those local diagnostics. It is
weaker and cleaner: the Qwen holdout cells show that temporal placement can
preserve answer accuracy at substantially lower effective fresh-frame budget on
the motion slices, while the paper reports C-VISION and C-PERSIST separately
when it needs measured wall-clock evidence.

\paragraph{Takeaway.}
Codec-like and pixel-domain signals do not need to be semantic oracles to be
useful. They need to be good enough freshness or uncertainty oracles that the
runtime buys new visual evidence in the right temporal neighborhood. The
stronger novelty-ranked and pixel-to-feature diagnostics remain useful
mechanism checks, but they are not needed for the headline frontier claim.

\section{Native-Rate Streaming as a Deployment Target}

Streaming state reuse is the natural scale-out target for anti-recomputation:
the model revisits nearly the same visual state while deciding when to keep,
shift, rebuild, or invalidate cached state. We do not claim a standardized
C-STREAM result yet. Instead, this section reports boundary evidence from a
26B-class scale-out evaluation on an M5 Max MacBook Pro with 128\,GB unified
memory: unsafe default cache paths, small-\(N\) after-warm prefix snapshot
positives, strong
low-FPS / screenshot / recency baselines, and denominator-safe paired-row
exports. The result is mixed, and that is the point: it shows which mechanisms
are plausible, which simple baselines are serious, and which cache paths are
unsafe.

Unlike sparse-sampled benchmark runs, the streaming lane tests cache lifetime:
when to keep state, shift it, rebuild it, or declare it stale. The tested
scale-out stack exercises motion-vector and residual side-channel staleness
signals, sectional cache shifts, and event/inflection triggers. Those
mechanisms are why the lane is not merely an application demo. Its claims stay
separately labeled because native streams add evaluation problems that
VideoMME, MVBench, and TOMATO under-test: query cadence, scroll/pan, live-world
updates, stale-cache failure, fixed-vision-budget perception rate, and cache
topology under mixed sliding-window/full-attention attention.

The local Qwen session/streaming bridge explains why this gate matters. The
dense-first-query reference returns aggregate first-query correctness to parity
and reaches 2.79$\times$ paired speedup, missing the preregistered 3.0$\times$
rescue floor. The later instrumented cache-reuse rerun preserves the same
aggregate boundary and reaches 3.02$\times$, but still misses the follow-up
fidelity target. Instrumentation also shows that follow-up vision pruning is
inactive in that family: follow-up image tokens are cache-served rather than
actively pruned. The right interpretation is dense-first-query admission plus
cache-state reuse, not a validated native-rate sparse-vision backend.

The 26B rows explain why cache-correctness is a gate rather than bureaucracy.
The default cross-turn cache path over mixed sliding-window and full-attention
layers fails the cache-correctness check: it can move fast while producing
divergent text and answer choices. A correctness guard that fully
refills rotating-cache layers passes, and the patched cache path passes the
full cache-correctness regression as well: 42/42 text-identical rows, zero choice or
correctness diffs, and four matched parse failures. That closes this
cache-correctness regression; it is not C-STREAM speedup. The patched path
refuses unsafe cross-turn trim and runs at roughly cold-dense wall-clock. An
after-warm prefix snapshot is
the positive scale-out follow-up row, but its denominator is narrow: the
reported 9.11$\times$ at 8f and
18.71$\times$ at 32f are per-follow-up medians after the prefix snapshot is
warm, not setup-inclusive session speedups. The 8f row has seven
multiple-choice prompts; five are parse-clean with 0/5 choice/correctness
diffs, and two have matched parse failures. The other 14 rows are open-ended
text-comparison rows, and 6/21 rows have text diffs. The expanded 32f row has seven
multiple-choice prompts with 0/7 choice/correctness diffs, plus 14 open-ended
text-comparison rows, 7/21 text diffs, and no parse failures. These are small-\(N\),
wrapper-specific rows, not proof that ordinary PromptCacheState is safe.

The same rows also answer a throughput question that streaming readers
naturally ask: what is the effective prompt-frame throughput on the warm
follow-up wall-clock denominator? The 8f prefix snapshot yields
27.02\,fps at the median, with 8/21 paired rows above 30\,fps. The expanded 32f
snapshot yields 54.68\,fps at the median, with 19/21 rows above 30\,fps and a
23.85--134.43\,fps range. Cold dense is about 3\,fps on the same rows.
Read this on the same after-warm prefix snapshot denominator: it is a throughput
view of those rows, not ingest throughput or setup-inclusive session speed. The
first query still pays full dense visual processing, the snapshot wrapper is
required for the safe path, and continuous capture, stale-cache invalidation,
and desktop-agent scroll/pan traces remain future gates.

The streaming baselines cut in the other direction. On 22 screen/UI events,
low-FPS dense uses four uniformly spaced frames and matches the fresh oracle on
17/22 events, beating screenshot polling with one query-time frame (13/22),
recency last-\(K\) with four recent frames (12/22), and the tested
event-window proxy with four event-window frames (13/22). That does not kill
C-STREAM; it clarifies the denominator. Fixed-evidence quality is not where the
tested proxy wins. The throughput view above shows why the native-rate denominator is
still worth measuring, but the evaluated bundle has not yet earned the stronger
streaming claim. Earlier sectional-scroll instrumentation points in the same
direction qualitatively, but the evaluated rows here are the evidence
surface for this paper.

The recovery path is therefore specific, not rhetorical. C-STREAM needs a
policy that wins on the axis native streams actually own: higher
perception-rate at the same vision budget while preserving event-window
answers. The tested proxy failed because its event windows were not enough to
beat a simple four-frame dense baseline; the default cache path failed because
mixed sliding-window/full-attention cache topology made cross-turn state
semantically unsafe; and post-vision-tower pruning failed as systems evidence
because it skipped work after the expensive stage had already run. A positive
C-STREAM result should combine topology-safe state reuse, an event/staleness
gate that beats low-FPS dense under matched event windows, and a stale-cache
failure case that shows when the runtime deliberately buys fresh evidence.

\begin{table}[H]
\centering
\caption{Evaluated scale-out bundle on an M5 Max MacBook Pro with 128\,GB unified memory. Rows are operational evidence; native mechanism quality, matched baselines, and speed remain open gates.}
\label{tab:scaleout-bundle}
\scriptsize
\renewcommand{\arraystretch}{\PaperTableStretch}
\begin{tabularx}{\linewidth}{@{}>{\raggedright\arraybackslash}p{0.22\linewidth} Y >{\raggedright\arraybackslash}p{0.25\linewidth}@{}}
\toprule
Probe & Result & Claim status \\
\midrule
Default 26B cache reuse & 16/42 text diffs; 2 choice diffs; 4 parse failures; unsafe cross-turn cache path & diagnostic boundary; rejected default path \\
Correctness guard & 0/42 text diffs under full-refill guard; 4 parse failures & correctness control; speed not evaluated \\
Patched cache-library closure & 0/42 text diffs; 0 choice diffs; 0 correctness diffs; 4 matched parse failures; 0.98$\times$ vs cold dense & correctness closure; not a speed path \\
26B prefix snapshot (8f) & after-warm follow-up 9.11$\times$; MC choice/correct 0/5 / 0/5; 2 matched parse failures; open-ended text diffs 6/21; 27.02 fps median (8/21 rows >=30 fps) & after-warm prefix snapshot; setup excluded; wrapper-specific \\
26B prefix snapshot (32f) & after-warm follow-up 18.71$\times$; MC choice/correct 0/7 / 0/7; open-ended text diffs 7/21; parse 0/21; 54.68 fps median (19/21 rows >=30 fps; 23.85--134.43 fps) & after-warm prefix snapshot; setup excluded; wrapper-specific \\
Fixed-evidence stream baselines & low-FPS dense 17/22 beats screenshot 13/22, event-window proxy 13/22, recency 12/22 & baseline pressure; throughput axis remains separate \\
Post-vision-tower hard prune & 8f 0.757$\times$, 32f 1.042$\times$; text diffs 10/10; timing-only descriptive prompts & overhead boundary; no correctness gate; not real sparse vision-tower speedup \\
Scale-out paired-row export & zero correctness delta on 1,937 logged rows; byte-identical raw-paired text on 513 rows & denominator-safe reporting \\
\bottomrule
\end{tabularx}
\end{table}

The scale-out bundle also adds two useful boundary checks. First, the paired-row
export keeps its denominators separate: zero correctness delta is logged on
1,937 sparse-sampled rows, while byte-identical raw-paired text is verified on
513 rows. Second, post-vision-tower hard pruning
on the 26B stack is a negative systems result: because the full vision tower has
already run, policy overhead can dominate, so the row is not evidence for real
sparse-ViT speedup. It is a reminder that the paper's C-CEILING arithmetic
assumes the accelerated stage is actually skipped and the policy overhead is
small.

\section{Discussion: Toward VLM-Native Media}

The experiments above are intentionally conservative: frozen models, and speed
claims only where the backend skipped timed work.
That conservatism should not hide the bigger implication. Existing media
pipelines already contain cheap temporal structure, but most video VLM stacks
consume the final RGB frames as if every frame were equally new. Anti-
recomputation asks a systems question before it asks for a new model: what did
the media pipeline already know, and why did the VLM pay to rediscover it?

\subsection{Unchanged Is Only The First Case}

Same-position reuse is the simplest regime: if a patch did not change, reuse
its visual evidence. But video codecs also exploit a richer fact: many things
change predictably. Objects translate, cameras pan, forward-driving cameras
produce structured expansion, and occlusions create localized new evidence.
This paper earns evidence mostly for static-to-medium-motion benchmark
regimes and for same-video follow-up reuse. It does \emph{not} yet prove
motion-compensated feature reuse, driving-specific routing, or sensor-fusion
caching. Those directions follow naturally from the failures and ceilings we
measured, so they belong in the paper's agenda rather than in the headline
claim.

\begin{table}[H]
\centering
\caption{Application regimes as evidence targets, not generic motivation. The
right policy depends on the redundancy structure and on what failure would
cost.}
\label{tab:application-regimes}
\footnotesize
\renewcommand{\arraystretch}{\PaperTableStretch}
\begin{tabularx}{\linewidth}{@{}
  >{\raggedright\arraybackslash}p{0.18\linewidth}
  >{\raggedright\arraybackslash}p{0.22\linewidth}
  >{\raggedright\arraybackslash}p{0.26\linewidth}
  >{\raggedright\arraybackslash}X
@{}}
\toprule
Regime & Redundancy structure & Candidate routing policy & Evidence status / risk \\
\midrule
Static cameras, factories, surveillance, stove/countertop & stable background,
localized entrants, rare decisive events & persistent visual memory plus
bounded-staleness refresh around moving regions & closest to tested evidence;
needs stronger application baselines and event-window recall \\
\addlinespace[0.25em]
Screen and UI video & exact-copy regions, glyph changes, scrolling, cursor
motion & screen-content path with exact-copy, OCR/glyph, and high-frequency
guards & planned specialization; VideoMME audit under-samples scroll/pan \\
\addlinespace[0.25em]
Forward driving & structured ego-motion, border entry, independently moving
objects & pose/flow-conditioned refresh with protected object and boundary
regions & future lane; should not be inferred from same-position reuse \\
\addlinespace[0.25em]
Egocentric, FPV, drones & jerky motion, parallax, frequent occlusion & global
stabilization or multi-reference cache before reuse & tested benchmark mix is
the wrong corpus for this claim \\
\addlinespace[0.25em]
Robotics / VLA & static workspace plus gripper/object/contact zones and hard
deadlines & protected ROIs, p95/p99 latency budgets, guarded dense fallback & future
systems lane; answer-stability QA is not a control-safety metric \\
\addlinespace[0.25em]
Games and HUD-heavy streams & stable HUD, structured camera regimes, repeated
interaction states & HUD anchors plus event-triggered world refresh & future
application lane; likely high reuse but unmeasured here \\
\bottomrule
\end{tabularx}
\end{table}

\paragraph{Robotics is a different contract.}
Robotics is not just another row in an application table. It changes the
optimization target from average answer stability to deadline-safe perception.
A plausible safety contract is protected reuse, not maximal reuse: test
policies that always refresh the gripper, manipulated object, contact boundary,
goal receptacle, newly entering regions, and any sensor-disagreement zone;
reuse aggressively elsewhere; and report p95/p99 latency, deadline misses,
fallback rate, and task success instead of only QA accuracy. We do not claim
that evidence here. We name it because it is the natural systems lane once
anti-recomputation leaves offline QA.

\subsection{From Video Streams To State Streams}

The scale-out streaming rows motivate this interface: they are not yet a
standardized C-STREAM result, but they show why a perception-rate state stream
is the right deployment abstraction to test.

The same logic extends beyond RGB video. A robot, vehicle, or factory cell may
carry multiple cameras, depth or ToF, IMU/odometry, event sensors, object
tracks, calibration, timestamps, and confidence estimates. These side channels
are not semantic-saliency oracles. They are cheaper refresh and uncertainty
oracles: RGB says what a region looks like; depth says whether geometry moved;
IMU says whether apparent motion is explained by the camera; event sensors
flag onset; multi-camera geometry distinguishes occlusion from disappearance.
The VLM-native interface we want is therefore not ``more frames,'' but a
synchronized state-update stream that tells the runtime where fresh visual
evidence is worth buying.

\subsection{Codec Signals Are Requirements Probes}

Classical codecs are hand-engineered prediction systems: predict a reference,
send motion and residual surprise, transform and quantize under a rate-
distortion objective, and make the result cheap enough for hardware. That
objective was designed mainly for human viewing. Standards work is already
moving toward machine consumption, including MPEG-AI Part 2 Video Coding for
Machines \cite{mpegvcm}, JPEG AI \cite{jpegai}, and event-data coding efforts
such as JPEG XE \cite{jpegxe}. Older object/layer media ideas, including
MPEG-4 Visual object coding \cite{mpeg4visual}, also show that rectangular RGB
frames were not the only conceivable abstraction. The analogy to object-based
audio such as Dolby Atmos \cite{dolbyatmos} is useful but only as an analogy:
the goal is not theatrical rendering, but a synchronized stream of
model-facing state updates.

Our role in that trajectory is narrower. We are not designing the future codec
here. We use VLM behavior to motivate candidate requirements for that future
interface:

\begin{table}[H]
\centering
\caption{How observed limitations motivate future media-interface requirements.
These are hypotheses and candidate design requirements, not measured claims in
this paper.}
\label{tab:media-requirements}
\small
\renewcommand{\arraystretch}{\PaperTableStretch}
\begin{tabularx}{\linewidth}{@{}
  >{\raggedright\arraybackslash}p{0.40\linewidth}
  >{\raggedright\arraybackslash}X
@{}}
\toprule
Observed limitation or boundary & Future VLM-native media should expose \\
\midrule
Mean block difference misses small but decisive events & residual
concentration, changed-pixel fraction, onset flags, and max-age state \\
Novelty-ranked dense frames can cluster budget in the wrong place & temporal
coverage constraints, event-window hints, and placement-aware scheduling \\
Same-position reuse weakens under camera motion & camera pose, global motion,
depth-aware flow, and boundary-entry flags \\
Predictably moving objects are not static but are not arbitrary either &
object IDs, masks, tracks, motion uncertainty, and multi-reference state \\
Human-compression artifacts can hide model-critical text or color & task-aware
text, chroma, palette, and high-frequency metadata \\
Sparse regions may be awkward for today's tensor kernels & tensor-friendly
active tiles, projector-consistent groups, and regular batched sparse layouts \\
RGB alone confuses lighting, geometry, and contact & synchronized depth/ToF,
event, IMU/odometry, timestamp, and confidence sidecars \\
\bottomrule
\end{tabularx}
\end{table}

\paragraph{Layered world-state media.}
One useful mental model is a machine-facing analogue of object/layer media:
static background memory, camera pose, object tracks and masks, motion
uncertainty, residual surprise, text/glyph events, event-camera onset,
depth/ToF changes, timestamps, and confidence. The model would render or
refresh dense visual evidence only where the state update requires it. This is
not a codec implementation; it is the interface pressure revealed by a
paper about paying for the wall again.

\paragraph{Beyond zero-change inference.}
The result is useful precisely because it is frozen-model and
training-free. It identifies where an existing model already tolerates reuse
and where it does not. The model-changing version should train around those
traces rather than ignore them: learned recache gates, codec-conditioned token
types, P-frame or delta encoders, and cache-state-aware attention can all use
the failure cases and routing logs produced by this paper as supervision.
There is also a training-compute hypothesis: if future VLMs learn from
state-update streams rather than dense RGB frame stacks, they may spend the
same training FLOPs on more temporal coverage, more queries per clip, or harder
event windows. We do not measure that here. The caveat is important:
post-vision-tower feature caching does not directly become an end-to-end training
speedup when the vision encoder is trainable, because cached features
approximate or break the gradient path. The training version likely needs
model-native delta encoders, differentiable refresh gates, or teacher/student
traces rather than dropping inference caches into pretraining unchanged.

\paragraph{Beyond video.}
The general pattern is not unique to pixels. Any expensive encoder over a
slowly changing state stream can, in principle, benefit from a cheap freshness
oracle: financial order books, industrial telemetry, scientific sensor arrays,
long medical monitoring streams, or logs from embodied agents. Video is the
right starting point because temporal redundancy is explicit, codecs already
extract motion and residual signals, and the cost of visual encoding is large
enough to matter. Other domains need their own state variables and failure
contracts; the transferable lesson is anti-recomputation under a measured
quality--compute frontier, not pixel differencing itself. The transfer rule is
strict: the domain needs slowly changing state, an oracle cheaper than the
encoder, expensive recomputation, and task-specific paired-failure metrics.
Without all four, anti-recomputation is just a metaphor.

\subsection{Open Experiments}

Several high-value brainstorm items remain deliberately outside the tested
evidence boundary.

The near-term research questions are reader-facing, not bookkeeping. Can Qwen
routing be converted from semantic substitution into a broad measured sparse
backend, and can sparse LM prefill join the timed-skip envelope? Does temporal
placement of fresh compute beat novelty magnitude under matched budgets? Can
saved compute buy broader temporal coverage rather than only lower latency? Can
answer-margin and paired-drift features predict when reuse will fail? Which
application baselines make deployment claims credible rather than anecdotal?
How far does paired stability extend when a user asks dialogue-like follow-up
questions about the same video, and what repair schedule prevents drift from
accumulating?

The longer-range agenda is the VLM-native media ladder: object/state delta
sidecars; sensor-fusion world-state streams with depth/ToF, event, pose, and
timestamp layers; compute-denial robustness; learned recache gates;
codec-conditioned token types; learned delta encoders; and hardware-aware
active-tile decode. These ideas follow from the measurements above, but they
remain future work unless they receive their own reproduced evidence.

The same anti-recomputation principle may also rotate into generative video
and audio. A video generator or world model could reuse latent hidden state
across denoising steps or autoregressive chunks when predicted motion and
residual uncertainty are low, then refresh around occlusion, entropy spikes, or
new-object events. Audio has a similar shape but a different trap: quiet is not
the same as irrelevant, so onset, phoneme boundaries, speaker changes, and
prosody matter more than energy alone. These are domain rotations, not evidence
claims in this paper.

The long-term version is simple to state: video for machines should become a
timeline of state updates, not only dense RGB frames. Static background,
camera motion, dynamic objects, text, events, depth, sensor confidence, and
time can all be first-class evidence. This paper is the first rung: a careful
training-free measurement of where anti-recomputation works today, where it
fails, and what future media systems should stop making the model rediscover.

\section{Limitations and Reproducibility}

The paper makes three kinds of claim, and each one breaks differently. For
first-pass pruning, VideoMME is the clean holdout anchor, while MVBench and
TOMATO keep timing caveats. Gemma supplies the cleanest measured sparse-vision
cell: 32f short skips timed vision-tower work at 1.316$\times$ first-query
speedup with no paired drift and no parse failures on 20 items. The broader
Gemma sweep has matched parse-failure caveats, and Qwen supplies the boundary:
its 8f keep-rate sweep follows the timing ceiling while failing fidelity, while
\(kr_V=0.85\) at 16f recovers to within aggregate-accuracy, format, and ceiling
gates at only 1.032$\times$. The claim is a measured sparse-vision envelope,
not a universal sparse backend.

For persistent-KV, the larger multipliers are narrower by construction. They
are \emph{after-ingest} claims: the first query still pays full freight, and
the gain appears when later questions reuse the same video state. Unrepaired 7B
reuse is clean through the tested 16f point and then enters a basin; 3B holds a
wider but weaker pre-basin plateau. Selective re-prefill repairs that failure
inside the tested envelope. Fixed \(K=1\) and adaptive repaired-cache
inheritance both show no observed paired drift on the 93-query breadth setting,
with adaptive reaching 14.90--35.92$\times$ same-class follow-up speedup. The
sample is still finite: the query-level rule-of-three bound is about 3.2\%,
and the session-level 0/31 view gives about 9.7\%. The many-turn stress checks
repeated-question cache horizon, not natural dialogue; the dense-answer-anchored
stress is stricter and exposes the speed/fidelity tradeoff, with fixed \(K=1\)
at 0/133 and aggressive policies at 6/133 drift.

\begin{table}[H]
\centering
\caption{Memory characterization across reported cells. The runtime uses an MLX allocation cap, and observed process working-set peaks still reached 13.6\,GB.}
\label{tab:memory-characterization}
\small
\renewcommand{\arraystretch}{\PaperTableStretch}
\begin{tabularx}{\linewidth}{@{}Y r r r Y@{}}
\toprule
Family & Cells & Max peak & Cells \(>10\)GB & Paper meaning \\
\midrule
After-ingest follow-up reuse & 45 & 13.61\,GB & 24 & observed envelope \\
Composition/admission boundary & 10 & 6.67\,GB & 0 & observed envelope \\
Measured sparse vision execution & 23 & 10.80\,GB & 4 & observed envelope \\
\bottomrule
\end{tabularx}
\end{table}

The related-work bar for this lane is also higher than ``we reused a KV
cache.'' Prefix caching, RadixAttention, CacheBlend-style repair, SparseVILA,
and HERMES all make parts of that sentence prior art. The manuscript's
claim is narrower: video-specific visual-tail repair, paired-drift accounting,
the cache-basin failure mode, and the repaired-cache inheritance rule on this
local stack. A matched head-to-head against the closest persistent-video or
visual-token sparsity baselines remains venue-hardening work rather than part
of the main claim.

For Qwen routing, the opposite caveat applies. The holdout evidence is clean
and useful, but it is still a quality--compute frontier result under a fixed
dense backend. It should not be inflated into a measured sparse-execution
speedup claim until the backend actually skips timed work. The same caution
applies to the Qwen VideoMME frame-count story: the dev-split 16f long-bucket
regression does not replicate on disjoint holdout, so the manuscript treats it
as a dev-split shape rather than a universal scaling law. The feature-drift
geometry itself reproduces on holdout at 8f and 16f; what failed to
transfer was the downstream long-bucket accuracy shape, not the drift signal.

The streaming/deployment lane remains the least harmonized evidence class, but
it names the practical regime where redundancy can become largest. The
scale-out bundle is mixed: default 26B cross-turn cache reuse is unsafe, an
after-warm prefix snapshot is positive only at small \(N\), fixed-evidence
stream baselines favor low-FPS dense over the tested event-window proxy, and
post-vision-tower hard pruning shows that policy overhead can erase any
expected win if vision work was not actually skipped.

The Qwen composition bridge is useful precisely because it is diagnosed rather
than promoted. Dense-first-query admission returns Q0 aggregate correctness to
parity,
but the three-query bridge still misses the rescue gate; the long-bucket
admission sweep falsifies every tested Q0 keep rate from 0.67 to 0.90. The
mechanistic boundary is sharper than aggregate loss: cache reuse preserves the
speed profile but leaves follow-up-state damage, while cache invalidation makes
follow-up V-pruning active, collapses speedup to 1.06$\times$, and still lands
on the same net aggregate loss through different drift sets. The cache-distance
probe supports that distinction, but it does not provide a deployable row-level
drift guard; the next proof step is causal intervention or independent state
features.

A related local codec-native bridge remains scientifically interesting, and
the codec-native planner-substitution results provide semantic
planner-substitution evidence: MAX-over-span sparse sampling is hard-falsified,
but the continuous-score redesign matches dense choices on VideoMME dev
all-duration \(n=30\) and later calibration-mode/source ablations are neutral
on the tested local slices. This remains a semantic planner result, not a
latency win; the extraction path is offline and not integrated into the timed
deployment path.

The tested benchmark mix also under-samples scroll and pan regimes. We measured
this directly on a 60-item VideoMME stratification across 8f/16f/32f and found
0/60 items above a relaxed \texttt{shifted\_fraction} \(\geq 0.30\) gate, with
max shifted fraction 0.125. That does not certify C-VISION on sustained
egomotion or scrolling content; it shows VideoMME is the wrong corpus for that
question. A scroll/pan characterization now requires EgoSchema, EPIC-Kitchens,
Ego4D, or a labeled synthetic set.

A dynamic-compute runtime also creates a robustness surface we have not yet
measured. Inputs can amplify apparent novelty and force the system to spend
more compute seeing: flicker, rolling-shutter bands, blinking UI, repetitive
local motion, scene-cut bursts, camera shake, and sensor desynchronization are
all future stressors. The right metrics are novelty amplification, latency
amplification, deadline miss rate, and fallback behavior. Until that suite
exists, the speedups in this paper are workload-regime results, not worst-case
latency guarantees.
The dual failure also matters: an input can hide a task-critical change below
the refresh trigger and cause stale reuse rather than excessive recompute. A
real robustness suite needs both denial-of-savings and denial-of-freshness
cases.

Simple dense baselines are also not strawmen. A 4f cold-dense baseline over the
same 57-session / 171-query VideoMME union is only 1.21$\times$ faster than 8f
and has ambiguous overall accuracy delta, but it is a real competitor in some
duration buckets and a hard short-bucket failure. That is why future C-STREAM
and C-VISION claims need matched low-FPS dense controls rather than only
same-method ablations.

Finally, equal aggregate accuracy is not the same as identical behavior. Gemma
MVBench preserves aggregate accuracy while changing 8/30 choices across four
motion subgroups, and the composition and selective-reprefill drift audits show
the same pattern in different forms. The paired-drift columns are part of the
claim.

The persistent-cache failures should also be read as attention-context or
cache-state failures, not as positional-encoding drift claims. The recurring
failure mode is not just stale pixels; it is incoherent internal state. Fresh
and cached mixtures must preserve cache topology, positions, and layer
semantics, or they can move fast while answering the wrong question. These
experiments identify when reused state changes later answers and show that
pruned key/value state can diverge strongly from dense. They do not yet provide a
general cache-validity guard or a unique row-level causal predictor.

Architecture exposure is another limitation and a design lesson. Qwen is the
high-stability local family for routing and persistent-cache follow-up reuse.
Gemma is the primary first-pass vision-pruning family and exposes a useful
vision-tower sparse path, but Gemma adaptive persistent-cache reuse remains
blocked by sliding-window / rotating-cache semantics in the tested MLX stack.
Future VLM runtimes should expose stable cache semantics, cache invalidation
controls, vision-tower keep masks, per-stage timing, token geometry, and
cache-state diagnostics as first-class interfaces. Those hooks are not just
engineering convenience; they decide which anti-recomputation claims can be
tested without risking silent cache corruption.

\begin{table}[H]
\centering
\caption{Architecture lessons from the tested evidence. This is not a model
leaderboard; it is a runtime-interface lesson about which reuse claims can be
tested cleanly.}
\label{tab:architecture-lessons}
\small
\renewcommand{\arraystretch}{\PaperTableStretch}
\begin{tabularx}{\linewidth}{@{}>{\raggedright\arraybackslash}p{0.18\linewidth} Y Y@{}}
\toprule
Family & Tested support & Design lesson \\
\midrule
Qwen2.5-VL-7B-Instruct & Stable routing and persistent follow-up reuse; strong
selective re-prefill evidence; measured sparse-vision configuration boundary &
Expose stable cache semantics, cache diagnostics, and frame-tail repair hooks. \\
Gemma 4-E4B & First-pass vision-pruning measurements; clean measured
sparse-vision 32f-short operating point plus broader zero-drift sweep with
format caveats & Expose keep masks and per-stage timing, but make
rotating/sliding cache behavior explicit. \\
Candidate C-STREAM & Native-rate streaming state reuse under a separate
protocol; evaluation bundle with cache-correctness, prefix snapshot, fixed-evidence
baselines, and paired-row export & Promote only after the native streaming
policy beats low-FPS/screenshot/recency baselines under the same quality
envelope and cache topology. \\
\bottomrule
\end{tabularx}
\end{table}

\section{Conclusion}

The paper began with a small absurdity: much of the video did not change, and
the model paid anyway. The measured answer is not one multiplier.
Same-video follow-up reuse is the big-number regime: after the first query
pays, selective re-prefill and repaired-cache inheritance give
14.90--35.92$\times$ follow-up speedup with no observed paired drift across
the 93-query breadth setting. Fresh-video pruning is smaller but real, and its
gains obey the stage-share ceiling. Streaming is the natural deployment target,
but the evaluated 26B-class bundle shows why topology-safe cache reuse and strong
low-FPS baselines are gates, not details.

The boundaries are the science. Aggregate accuracy can hide answer drift.
Semantic substitution is not timed skipped work. Component wins do not multiply
just because the table numbers look good next to each other.

The larger direction is simple: VLMs should not have to reconstruct world state
from repeated frames. Future media streams for machines should expose change,
motion, uncertainty, objects, text, sensor time, and active tiles directly.
Anti-recomputation is one empirical step from video-as-frames toward
video-as-world-state updates.

\appendix
\section{Source Traceability and Provenance}
\label{app:source-traceability}

The public repository is \url{https://github.com/jfbastien/VLMaxxing}.
The arXiv source archive includes the LaTeX source and generated figures and
tables used to render this version. This PDF was built from commit
\texttt{\PrimaryRepoSHA} dated \PrimaryRepoCommitDate.

\ifshowclaimmanifest
The evidence labels are shorthand, not a separate claim taxonomy. \emph{Local
clean} means checked local rows that pass the relevant quality and timing gates.
\emph{Local advisory} means checked local rows whose timing attribution or
strict quality gate is imperfect. \emph{Local boundary}, \emph{bounded},
\emph{negative}, or \emph{sanity} rows define limits, failures, or narrow
operating envelopes. \emph{Scale-out bounded / mixed} means checked 26B-class
bundle rows with raw paired rows and matched baselines, but not C-STREAM
closure. Hypotheses and future-work items do not support the main quantitative
claims.

\begingroup
\scriptsize
\setlength{\tabcolsep}{2pt}
\renewcommand{\arraystretch}{1.28}
\begin{longtable}{@{}
  >{\raggedright\arraybackslash}p{0.14\linewidth}
  >{\raggedright\arraybackslash}p{0.39\linewidth}
  >{\raggedright\arraybackslash}p{0.17\linewidth}
  >{\raggedright\arraybackslash}p{0.25\linewidth}
@{}}
\caption{Claim/protocol/evidence manifest for the main quantitative assets.}
\label{tab:claim-protocol-manifest}\\
\toprule
Asset & Claim / protocol & Evidence class & Source / hardware \\
\midrule
\endfirsthead
\caption[]{Claim/protocol/evidence manifest for the main quantitative assets (continued).}\\
\toprule
Asset & Claim / protocol & Evidence class & Source / hardware \\
\midrule
\endhead
\midrule
\multicolumn{4}{r}{Continued on next page} \\
\endfoot
\bottomrule
\endlastfoot
Headline table & regime headline cells; first-pass vision-pruning holdout, measured sparse-vision envelope, and after-ingest follow-up & local clean + local advisory & checked C-VISION holdout, measured sparse-vision, and C-PERSIST repair artifacts; Gemma/Qwen local stacks, MLX / M3 Air 16 GB \\
C-CEILING figure & C-CEILING / C-VISION scatter-back validation; first-pass holdout, n=60 stacked-regime audit, matched cross-arch probe, and measured sparse-vision boundary & local clean + local advisory + local boundary & ceiling data JSON; local Gemma and Qwen stacks plus checked-in TOMATO rerun and measured sparse-vision artifacts \\
Measured sparse-vision tables & real timed vision-tower work skipped under Gemma and Qwen & local bounded systems evidence & Gemma 8/16/32f scaling, Qwen 16f keep-rate sweep, and Qwen 8f boundary artifacts \\
C-PERSIST figure & C-PERSIST tested envelope; after-ingest same-video follow-up & local & Qwen2.5-VL 7B/3B-4bit follow-up reuse artifacts, MLX / M3 Air 16 GB \\
C-PERSIST repair & 20f/32f same-video selective re-prefill & local bounded recovery & fixed and adaptive selective re-prefill artifacts plus stage-timing attribution; Qwen2.5-VL-7B-Instruct-4bit, VideoMME short/medium/long + 32f short \\
C-PERSIST sampler sweep & adaptive short-cell practical temperature robustness & local robustness evidence & sampler temperature and seed-sweep artifacts; Qwen2.5-VL-7B-Instruct-4bit, VideoMME short slice \\
C-PERSIST many-turn stress & 50-turn repeated-question cache horizon & local bounded stress & many-turn C-PERSIST artifacts; Qwen2.5-VL-7B-Instruct-4bit, seven VideoMME short videos, 20f, stateless question cycle \\
C-PERSIST dense-answer-anchored stress & content-conditioned follow-up prompt variation & local bounded stress & dense-answer-anchored C-PERSIST artifacts; Qwen2.5-VL-7B-Instruct-4bit, seven VideoMME short videos, 20f, identical dense/cached prompt hashes \\
C-PERSIST economics & setup-inclusive same-video session speedup after paying the first-query cost & local serving arithmetic & setup-inclusive table built from raw persistent-KV timing artifacts; Qwen2.5-VL 7B/3B-4bit frame-scaling cells \\
Routing frontier & sparse routing under dense backend & local clean & checked routing artifacts and generated routing snapshot; Qwen2.5-VL-7B-Instruct-4bit, MLX \\
Gemma routing boundary & second-architecture semantic substitution & local boundary & Gemma 4-E4B-4bit routing-boundary artifacts; TOMATO/MVBench motion holdouts \\
Qwen stacked boundary & session/streaming composition plus cache-boundary attribution & local negative / boundary & Qwen composition, cache-invalidation, and attribution artifacts; Qwen2.5-VL-7B-Instruct-4bit, VideoMME dev+holdout \\
Cache-distance probe & fp32 key/value cache-distance audit for composition boundary & local mechanism instrumentation & 20-row balanced drift-class cache-distance artifact; native fp16 cosine retained as overflow diagnostic, fp32 cosine is the reported metric \\
Low-FPS dense baseline & 4f dense versus 8f dense on composition corpus & local baseline / ambiguous & low-FPS dense artifacts; Qwen2.5-VL-7B-Instruct-4bit, 57 sessions / 171 queries \\
Qwen 8f keep-rate sweep & C-CEILING timing validation under failed-fidelity sparse vision & local timing boundary & Qwen 8f keep-rate artifacts at \(kr_V=0.25/0.50/0.75\); timing matches prediction while fidelity fails \\
Random-keep sanity check & structured Qwen scorer versus uniform random keep at matched keep-rate & local sanity baseline & Qwen 8f VideoMME dev30 dense, magnitude-score, and four uniform-random seeds \\
Paired drift table & paired choice/correctness drift summary & local analysis & paired-drift summary built from composition, Gemma routing-boundary, and persistent-cache artifacts \\
Stability predictor scouts & dense margin and richer features are insufficient as deployable drift guards & local negative / future guard feature & stability-predictor artifacts; 228 scored Qwen paired rows \\
Memory envelope & process working-set characterization across reported families & local reproducibility evidence & memory characterization artifacts; 78 cells across follow-up reuse, admission/composition, and measured sparse vision \\
Method / limitations & codec-native planner bridge; semantic planner substitution & local semantic evidence & Codec-native planner-substitution artifacts; Qwen2.5-VL-7B-Instruct-4bit, VideoMME dev/short slices; offline extraction, not latency \\
Native-rate streaming bundle & 26B cache-correctness, prefix snapshot, fixed-evidence stream baselines, paired-row export, and post-vision-tower hard-prune boundary & scale-out bounded / mixed & Checked M5 Max MacBook Pro 128\,GB scale-out bundle, r2 correctness guard, patched-library closure, and expanded 32f n=21 prefix snapshot; earlier throughput-only evidence is advisory because raw paired rows are outside the paper's checked evidence set \\
\end{longtable}
\endgroup
\fi

\clearpage
\section{Real-Clip Routing-Budget Visualization}
\label{app:real-clip-planner}

\begin{figure}[H]
  \centering
  \IfFileExists{generated/figures/fig1_appendix_broadened/qwen_routing_budget_visualized.pdf}{
    \includegraphics[width=0.98\linewidth]{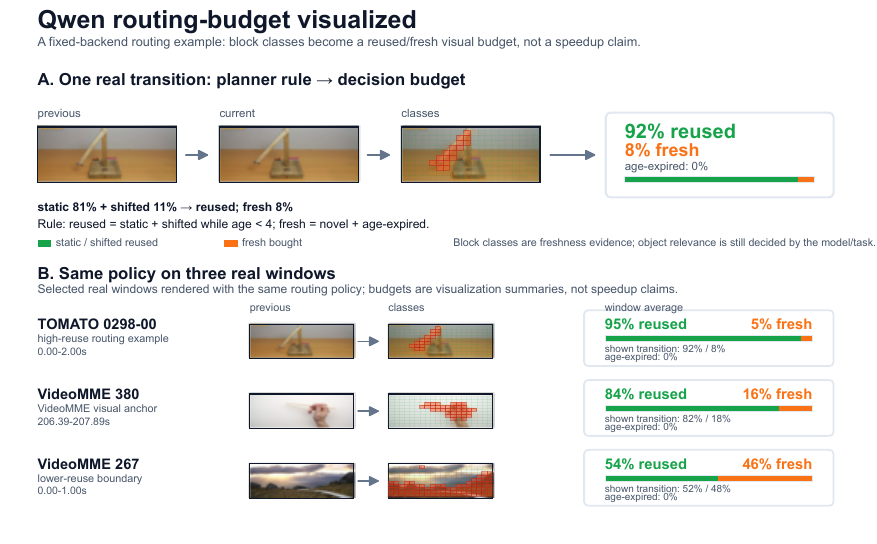}
  }{
    \fbox{\parbox{0.9\linewidth}{Real-clip routing-budget visualization unavailable in this source snapshot.}}
  }
  \caption{Qwen routing-budget visualization on real clips. Panel A overlays
  static, shifted, and fresh block classes for one transition, then reports
  reused versus fresh active-region budget; static/shifted blocks are reused
  until age four, while novel or expired blocks are fresh. Panel B applies the
  same policy to three short windows: a high-reuse TOMATO example, a VideoMME
  visual anchor, and a lower-reuse boundary. These rows visualize the routing
  budget, not additional benchmark-frontier points or speedup claims.}
  \label{fig:qwen-routing-budget-visualized}
\end{figure}

Figure~\ref{fig:qwen-routing-budget-visualized} visualizes the real-clip
budget behind the Qwen routing frontier in Section~\ref{sec:routing-mechanism}.
That result is a fixed-backend quality--compute claim: the reported policy
preserves answer quality while spending fewer effective fresh frames. The
figure shows how local active-region blocks are classified as reusable or fresh
before they are summarized into that effective visual budget. It is not
C-PERSIST timing, not C-VISION sparse execution, and not a semantic-saliency
map. Static and shifted blocks are treated as reusable until the age bound
requires refresh; fresh blocks are newly bought visual evidence.

\end{document}